\documentclass[11pt,english]{article} 
\usepackage[T1]{fontenc}
\usepackage[a4paper,margin=1.0in]{geometry}
\usepackage{times}   
\usepackage{helvet}   
\usepackage{courier}  
\usepackage{url}       
\usepackage{authblk}
\usepackage{enumitem}
\usepackage{amsmath}
\usepackage{graphicx}
\usepackage{caption}
\usepackage{wrapfig}
\usepackage{subcaption}
\usepackage{algorithm}
\usepackage{algorithmic}
\usepackage{amssymb,amsmath,amsthm}
\usepackage{multicol}
\usepackage{bbm}
\usepackage{microtype}
\usepackage{booktabs} 
\usepackage{multirow}
\usepackage{inputenc}
\usepackage{babel}
\usepackage[title]{appendix}
\usepackage{pifont}
\usepackage{booktabs} 
\usepackage[table, x11names]{xcolor}
\graphicspath{{image/}}

\usepackage{makecell}
\usepackage{tcolorbox}
\usepackage{wrapfig}

\usepackage{pgf,tikz,pgfplots}
\pgfplotsset{compat=1.14}
\usepackage{mathrsfs}
\usetikzlibrary{arrows}
\usetikzlibrary{arrows.meta, positioning, quotes}

\newcommand{\D}{\mathcal{D}}
\newcommand{\Proba}{\mathbb{P}}

\newcommand{\R}{\mathbb{R}}
\newcommand{\N}{\mathbb{N}}
\DeclareMathOperator*{\E}{\mathbb{E}}

\newcommand{\calX}{\mathcal{X}}
\newcommand{\calA}{\mathcal{A}}
\newcommand{\calY}{\mathcal{Y}}

\newcommand{\calS}{\mathcal{S}}
\newcommand{\calT}{\mathcal{T}}
\newcommand{\calM}{\mathcal{M}}
\newcommand{\calH}{\mathcal{H}}
\newcommand{\RN}[1]{%
  \textup{\uppercase\expandafter{\romannumeral#1}}%
}

\newcommand\independent{\protect\mathpalette{\protect\independenT}{\perp}}
\def\independenT#1#2{\mathrel{\rlap{$#1#2$}\mkern2mu{#1#2}}}

\newtheorem{theorem}{Theorem}
\newtheorem{lemma}{Lemma}
\newtheorem{corollary}{Corollary}

\theoremstyle{definition}

\title{Beyond $\calH$-Divergence: \\
Domain Adaptation Theory With Jensen-Shannon Divergence}
\author[1]{Changjian Shui\thanks{changjian.shui.1@ulaval.ca}}
\author[1]{Qi Chen}
\author[2]{Jun Wen}
\author[1]{Fan Zhou}
\author[1,3]{Christian Gagn\'e}
\author[4,5]{Boyu Wang}
\affil[1]{Universit\'e Laval}
\affil[2]{Zhejiang University}
\affil[3]{Mila, Canada CIFAR AI Chair}
\affil[4]{University of Western Ontario}
\affil[5]{Vector Institute}
\date{}
\graphicspath {{image/}}

\begin{document}
\maketitle
\begin{abstract}
\noindent We reveal the incoherence between the widely-adopted empirical domain adversarial training and its generally-assumed theoretical counterpart based on $\calH$-divergence. 
Concretely, we find that $\calH$-divergence is not equivalent to Jensen-Shannon divergence, the optimization objective in domain adversarial training. To this end, we establish a new theoretical framework by directly proving the upper and lower target risk bounds based on joint distributional Jensen-Shannon divergence. We further derive bi-directional upper bounds for marginal and conditional shifts. Our framework exhibits inherent flexibilities for different transfer learning problems, which is usable for various scenarios where $\calH$-divergence-based theory fails to adapt. From an algorithmic perspective, our theory enables a generic guideline unifying principles of semantic conditional matching, feature marginal matching, and label marginal shift correction. We employ algorithms for each principle and empirically validate the benefits of our framework on real datasets.
\end{abstract}

\section{Introduction}
Domain adaptation (DA)~\cite{pan2009survey} is commonly faced by machine learning practitioners, when the model is trained on a fixed source but is used for a slightly different target. To alleviate the performance degradation caused by such a distributional shift, many approaches have been developed in various fields such as computer vision~\cite{ganin2016domain}, natural language processing~\cite{guo2020multi}, and biomedical engineering~\cite{li2019target}. 

DA theory is crucial to the fundamental understanding and practical development of relevant algorithms. Conventionally, such theoretical guarantees were typically established on the notion of $\calH$-divergence~\cite{ben2007analysis,ben2010theory} and its subsequent variants~\cite{redko2020survey}, where it requires a small $\calH$-divergence between source-target and a small joint risk. In the context of representation learning, this quantity is minimized via the well-known domain adversarial training~\cite{ganin2016domain,long2015learning,tzeng2017adversarial}, which is a stimulating topic in current research. 

However, $\calH$-divergence theory itself is rather limited in many scenarios such as analyzing conditional shifts and understanding open set DA \cite{panareda2017open,cao2018partial,you2019universal}, which are commonly encountered by practitioners. In spite of some empirical success, the lack of rigorous theoretical analysis hampers its further advancement.
It has been noted that the inherent principle of domain adversarial training is to minimize the Jensen-Shannon divergence~\cite{goodfellow2014generative,nowozin2016f} of the source-target marginal distribution. Therefore, a DA theory established directly on the Jensen-Shannon divergence would provide a thorough understanding, and help overcome the limitations imposed by the use of $\calH$-divergence. 

In this work, we build a complete DA theoretical framework \emph{directly} based on Jensen-Shannon divergence. Indeed, we reveal that $\calH$-divergence is \textbf{not} consistent with the Jensen-Shannon divergence. Then we establish the upper bound of target risk is determined by the source error and the Jensen-Shannon divergence of two joint distribution (Sec.~\ref{subsec:joint_bound}). Moreover, we derive the upper bounds of bi-directional shifts (Sec.~\ref{subsec:two_directional}), including (a) Feature Marginal Shift $(\calT(x)\neq\calS(x))$ and Label Conditional Shift $(\calT(y|x)\neq\calS(y|x))$; (b) Label Marginal Shift $(\calT(y)\neq\calS(y))$ and Semantic (Feature) Conditional Shift $(\calT(x|y)\neq\calS(x|y))$. The theory provides a unified understanding of domain shifts, with cofeature shift and label shift being its special cases, which can provide intriguing theoretic insights and effective practice guidelines:

\textbf{Theoretical Insights:}\quad Jensen-Shannon divergence enables us to analyze the factors of label space that influence the transfer procedure, which remains elusive in the $\calH$-divergence. Specifically, ($\RN{1}$) we reveal that the intrinsic error of learning target-domain is controlled by the label-space size, source domain intrinsic error and the similarity of two domains (Sec.~\ref{subsec:cond}). ($\RN{2}$)  we also reveal why transfer learning is challenging if the label space of source and target are not identical (a.k.a. open set DA). We formally show that a smaller overlap over the label space leads to a more difficult transfer (Sec.~\ref{subsec:universal_da}).

\textbf{Practical Implications:}\quad  
Our theory motivates new DA practice for representation learning, which is missing in $\calH$-divergence. More concretely, we propose unified principles to control the target risk (Sec.~\ref{sec:semantic}):
($\RN{1}$) re-weighted semantic conditional matching, to control the feature conditional shift $D_{\text{JS}}(\calT(x|y)\|\calS(x|y))$; ($\RN{2}$) label marginal shift correction, as the way to eliminate the label marginal shift $D_{\text{JS}}(\calT(y)\|\calS(y))$; ($\RN{3}$) constraining the feature marginal shift, an approach to prevent poor target pseudo label predictions (i.e. predicted labels), a common phenomena that can lead to negative transfer in semantic conditional matching. The proposed guideline enables us to select existing algorithm for each principle. The empirical results on the real datasets verify the benefits of unified principles, compared with merely one 
or two of them (Sec.~\ref{sec:empirical_results}).

\section{$\calH$-Divergence based DA Theory}\label{sec:JStheory}
Supposing we have the source distribution $\calS$ and target distribution $\calT$ over the input and output space $\calX\times\calY$. According to \cite{ben2007analysis,ben2010theory}, if the data is generated by a marginal distribution and underlying labeling function pair $(\D,h^{\star})$, then the upper bound of target risk error w.r.t. $\forall h\in\calH$ is:
\begin{equation}
    R_{\calT}(h) \leq R_{\calS}(h) + 
    d_{\calH}(\calT(x),\calS(x)) + \beta,
    \label{eq:h_div}
\end{equation}
where $R_{\D}(h) = \E_{x\sim\D}|h(x)-h^{\star}(x)|$,  $d_{\calH}$ denotes the $\calH$-divergence for measuring the marginal distribution similarities and $\beta$ is the optimal joint risk over the two domains. 

As pointed out in \cite{ben2007analysis}, it is generally impossible to exactly estimate the $\calH$-divergence. Hence, this measure is approximated as a binary classification task where we are discriminating the source and the target samples. More specifically, the $\calH$-divergence is approximated by distance $d_{\calA} = 2(1-2\epsilon)$, with $\epsilon$ corresponding to the discrimination generalization error. Inspired by this intuition, \cite{ganin2016domain} and subsequent approaches empirically adopted adversarial loss \cite{goodfellow2014generative} between the domain classifier $d$ and feature extractor function $g$ in the context of representation learning:

\begin{equation}
\min_{g}\max_{d}\quad\E_{x_s\sim\calS(x)} \log(d \circ g (x_s)) +  \E_{x_t\sim \calT(x)} \log(1- d \circ g (x_t)),
\label{eq: gans_loss}
\end{equation}
\noindent where Eq.~(\ref{eq: gans_loss}) is the dual term of Jensen-Shannon divergence \cite{nowozin2016f}, i.e. $\min_{g}D_{\text{JS}}(\calS(g(x))\|\calT(g(x)))$.

\subsection{Jensen-Shannon Divergence is NOT the Proxy of $\calH$-Divergence}
\begin{figure}[htbp]
   \begin{subfigure}{0.4\textwidth}
   \centering
   \begin{tikzpicture}[x=0.5cm,y=1cm]
    \draw [line width=1pt] (2,4)-- (13.5,4);
    \draw (0.8,4) node {$\calT(x)$};
    \draw [line width=1pt] (2,3)-- (13.5,3);
    \draw (0.8,3) node {$\calS(x)$};
    \begin{scriptsize}
        \draw [fill=black] (2,4) circle (2.5pt);
        \draw[color=black] (2,4.5) node {\normalsize 0};
        \draw [fill=black] (4,4) circle (2.5pt);
        \draw[color=black] (4,4.5) node {\normalsize $2\xi$};
        \draw [fill=black] (6,4) circle (2.5pt);
        \draw[color=black] (6,4.5) node {\normalsize $4\xi$};
        \draw [fill=black] (8,4) circle (2.5pt);
        \draw[color=black] (8,4.5) node {\normalsize $6\xi$};
        \draw [fill=black] (10,4) circle (2.5pt);
        \draw[color=black] (10,4.5) node {\normalsize $8\xi$};
        \draw [fill=black] (12,4) circle (2.5pt);
        \draw[color=black] (12,4.5) node {\normalsize $10\xi$};
        \draw [fill=black] (3,3) circle (2.5pt);
        \draw[color=black] (3,3.5) node {\normalsize $\xi$};
        \draw [fill=black] (5,3) circle (2.5pt);
        \draw[color=black] (5,3.5) node {\normalsize $3\xi$};
        \draw [fill=black] (7,3) circle (2.5pt);
        \draw[color=black] (7,3.5) node {\normalsize $5\xi$};
        \draw [fill=black] (9,3) circle (2.5pt);
        \draw[color=black] (9,3.5) node {\normalsize $7\xi$};
        \draw [fill=black] (11,3) circle (2.5pt);
        \draw[color=black] (11,3.5) node {\normalsize $9\xi$};
        \draw [fill=black] (13,3) circle (2.5pt);
        \draw[color=black] (13,3.5) node {\normalsize $11\xi$};
    \end{scriptsize}
\end{tikzpicture}
\caption{} 
\end{subfigure}
\hfill
\begin{subfigure}{0.4\textwidth}
   \centering
    \begin{tikzpicture}[x=0.65cm,y=1.0cm]
        \draw [line width=1pt] (0,4)-- (8,4);
        \draw (0,4.5) node {$\calS(x)$};
        \draw (0,5)   node {$\calT(x)$};
        \begin{scriptsize}
            \draw [fill=black] (2,4) circle (2.5pt);
            \draw[color=black] (2,3.5) node {\normalsize 1};
            \draw[color=black] (2,4.5) node {\normalsize $1/3$};
            \draw[color=black] (2,5) node   {\normalsize $1/4$};
            \draw [fill=black] (4,4) circle (2.5pt);
            \draw[color=black] (4,3.5) node {\normalsize 2};
            \draw[color=black] (4,4.5) node {\normalsize $1/3$};
            \draw[color=black] (4,5) node   {\normalsize $1/2$};
            \draw [fill=black] (6,4) circle (2.5pt);
            \draw[color=black] (6,3.5) node {\normalsize 3};
            \draw[color=black] (6,4.5) node {\normalsize $1/3$};
            \draw[color=black] (6,5) node   {\normalsize $1/4$};
        \end{scriptsize}
    \end{tikzpicture}
\caption{}
\end{subfigure}
\caption{$D_{\text{JS}}(\calT(x)\|\calS(x))$ is not the approximation of $d_{\calH}(\calS(x),\calT(x))$: (a) for two uniform distributions with different supports, there exists $d_{\calH}(\calT(x),\calS(x))\ll D_{\text{JS}}(\calT(x)\|\calS(x))$ if $0<\xi\ll 1$; (b) while for two distributions with different probability mass, there exists $D_{\text{JS}}(\calT(x)\|\calS(x))< d_{\calH}(\calT(x),\calS(x))$}
\label{fig:counter_example}
\end{figure}
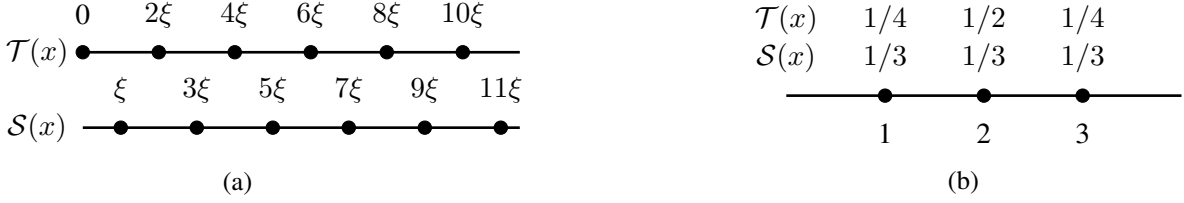

From Eq.~(\ref{eq: gans_loss}), domain adversarial training can be viewed as learning a representation to minimize the Jensen-Shannon divergence.  
However a $D_{\text{JS}}$ is not equivalent to $d_{\calH}$ in Eq.~(\ref{eq:h_div}). We find these two metrics can be very different and present two counterexamples to illustrate it, shown in Fig.~\ref{fig:counter_example}.

For the sake of simplicity, we design all the examples over one dimensional space and use the threshold functions $\calH=\{h_t: t\in\R\}$ as the hypothesis
class. That is, for any $t\in\R$, the threshold function is defined by $h_t(x)=1$ for $x<t$ and $h_t(x)=0$ otherwise.

\textbf{Counterexample 1}\quad  We adopt the example of \cite{ben2010impossibility}, showed in Fig.\ref{fig:counter_example}(a), with a small fixed $\xi\in(0,1)$. Let the target $\calT(x)$ be the uniform distribution over $\{2k\xi: k\in\N, 2k\xi\leq 1\}$ and the source $\calS(x)$ be the uniform distribution over $\{(2k+1)\xi: k\in\N, (2k+1)\xi\leq 1\}$.  We can compute $d_{\calH}(\calT(x),\calS(x))=\xi$ while $D_{\text{JS}}(\calT(x)\|\calS(x))=1$ since the two distributions have \emph{disjoint} supports. Then $d_{\calH}(\calT(x),\calS(x)) \ll D_{\text{JS}}(\calT(x)\|\calS(x))$ when $\xi\ll 1$, indicating a \emph{small $\calH$-divergence can correspond to a very large Jensen-Shannon divergence}. 

\textbf{Counterexample 2}\quad Fig.~\ref{fig:counter_example} (b) further illustrates that Jensen-Shannon divergence is even \emph{not} the upper bound of $\calH$-divergence. We assume the source $\calS(x)$ be the uniform distribution over $\{1,2,3\}$ and let the target $\calT(x)$ be the distribution on the same support with different probability mass $\{\calT(x=1)=1/4,\calT(x=2)=1/2,\calT(x=3)=1/4\}$. Then Jensen-Shannon divergence can be even smaller than $\calH$-divergence: $D_{\text{JS}}(\calT(x)\|\calS(x))< d_{\calH}(\calT(x),\calS(x))$. 

Due to these differences, $\calH$-divergence is \emph{not a proper} theoretical tool of analyzing the practice that minimizes the Jensen-Shannon divergence (e.g. domain adversarial training and its variants).

\section{DA Theory with Jensen-Shannon Divergence and Theoretical Insights}\label{sec: theoretical benefits}

\subsection{Upper and Lower Risk Bound}\label{subsec:joint_bound}
Slightly different from the settings in \cite{ben2010theory}, we assume the data $(x,y)\in\calX\times\calY$ is generated from a \emph{joint} distribution $\D$ and denote the hypothesis and loss function as $h:~\calX\times\calY\to \R$ and $L:~ \R\to\R$,
where the hypothesis $h\in\calH$ actually outputs a confidence score of an observation $(x,y)$. We also denote $R_{\D}(h)$ the expected risk w.r.t. distribution $\D$: $R_{\D}(h) = \E_{(x,y)\sim\D}~L(h(x,y))$.

\begin{theorem}[Upper Bound]\label{theorem: upper bound}
Supposing the prediction loss $L$ is bounded within an interval $G$: $G = \max(L) -\min(L)$, then for all the hypothesis $h$ the expected risk w.r.t. the target domain can be upper bounded by:
\begin{equation*}
    R_{\calT}(h) \leq R_{\calS}(h) + \frac{G}{\sqrt{2}}\sqrt{D_{\text{JS}}(\calT\|\calS)},
\end{equation*}
where $D_{\text{JS}}(\calT\|\calS)=\frac{1}{2}[D_{\text{KL}}(\calT\|\calM) +D_{\text{KL}}(\calT\|\calM)]$ with $\calM = \frac{1}{2}(\calT+\calS)$ is the Jensen-Shannon divergence between the joint distribution $\calS(x,y)$ and $\calT(x,y)$.
\end{theorem}

The proposed upper bound can be further extended to the \emph{unbounded loss} with sub-Gaussian or sub-Gamma property \cite{boucheron2013concentration} (see Appendix) and seamlessly connects the well known assumptions in DA. When the \emph{Cofeature shift} assumption holds ($\calT(x)\neq \calS(x)$, $\calT(y|x)=\calS(y|x)$), the upper bound can be expressed as  $R_{\calT}(h) \leq R_{\calS}(h) + \frac{G}{\sqrt{2}}\sqrt{D_{\text{JS}}(\calT(x)\|\calS(x))}$.  Besides, when the \emph{Label shift} assumption holds ($\calT(y)\neq \calS(y)$, $\calT(x|y)=\calS(x|y)$), the upper bound can be alternatively expressed as $R_{\calT}(h) \leq R_{\calS}(h) + \frac{G}{\sqrt{2}}\sqrt{D_{\text{JS}}(\calT(y)\|\calS(y))}$.

\begin{theorem}[Lower Bound]\label{theorem:lower_bound}
If we assume the loss $L$ as zero-one binary loss, then for any $h$, we can prove the target risk is lower bounded by:
\begin{equation*}
    R_{\calT}(h) \geq R_{\calS}(h) - \sqrt{D_{\text{JS}}(\calT\|\calS)}.
\end{equation*}
\end{theorem}

The lower bound provides the insights of the \emph{easy transfer} \cite{hanneke2019value} scenario: learning the target domain can be easier than the source domain, and the gap is controlled (smaller than) by their distribution distance.  For example, if we assume $R_{\calS}(h) = 0.2$, $D_{\text{JS}}(\calT\|\calS) = 2\times 10^{-4}$, then the target risk is also bounded: $R_{\calT}(h)\in [0.186,0.21]$. This indicates $R_{\calT}(h)$ can be smaller than $R_{\calS}(h)$ but not an arbitrary large gap. 

\subsection{Bi-Directional Marginal/Conditional Shifts}\label{subsec:two_directional}
We can decompose the joint Jensen-Shannon divergence into bi-directional marginal and conditional shift upper bound, according to the information theoretical chain rule \cite{Polyanskiy2019}.
\begin{corollary}\label{corollary:conditional_shift}
The upper bound in Theorem 1 can be further decomposed as:
\begin{equation}
\begin{split}
      R_{\calT}(h) \leq & \quad R_{\calS}(h) + \frac{G}{\sqrt{2}}\underbrace{\sqrt{D_{\mathrm{JS}}(\calT(x)\|\calS(x))}}_{\text{Feature Marginal Shift}} \\
      & + \frac{G}{\sqrt{2}}\underbrace{\sqrt{\E_{x\sim\calT(x)} D_{\mathrm{JS}}(\calT(y|x)\|\calS(y|x)) + \E_{x\sim\calS(x)} D_{\mathrm{JS}}(\calT(y|x)\|\calS(y|x))}}_{\text{Label Conditional Shift}}
\end{split}
\label{eq:x_cond}
\end{equation}
\begin{equation}
\begin{split}
      R_{\calT}(h) \leq & \quad R_{\calS}(h) + \frac{G}{\sqrt{2}}\underbrace{\sqrt{D_{\mathrm{JS}}(\calT(y)\|\calS(y))}}_{\text{Label Marginal Shift}} \\
      & + \frac{G}{\sqrt{2}}\underbrace{\sqrt{\E_{y\sim\calT(y)} D_{\mathrm{JS}}(\calT(x|y)\|\calS(x|y)) + \E_{y\sim\calS(y)} D_{\mathrm{JS}}(\calT(x|y)\|\calS(x|y))}}_{\text{Semantic (Feature) Conditional Shift}}
\end{split}
\label{eq:y_cond}
\end{equation}
\label{corollary_1}
\end{corollary}
In particular, Eq.~(\ref{eq:y_cond}) provides an alternative direction for understanding DA. The target risk bound is alternatively controlled by the label marginal shift and the semantic (feature) conditional distribution shift. Generally the source and target label marginal distribution, as well as the semantic (feature) conditional distributions are both different. For example, in the classification of different digits dataset (e.g. MNIST, USPS), when conditioning on the certain digit $Y=y$, it is clear that $\calS(x|Y=y)\neq \calT(x|Y=y)$, indicating the necessity of considering semantic information in DA.

\subsection{Theoretical Applications}
One fundamental challenge in DA is to discover the relations and inherent properties of learning tasks, that ensure a successful transfer  \cite{ben2010impossibility}.
Jensen-Shannon divergence enables us to analyze the factor of label space that influences the transfer procedure, illustrated in two concrete scenarios.

\subsubsection{Application $\RN{1}$: Target Intrinsic Error In DA} \label{subsec:cond}

To characterize the inherent difficulty in learning a task, we adopt the conditional entropy $H(Y_{\D}|X_{\D}) = \E_{x\sim\D(x)} H(Y|X=x)$ as the intrinsic error, an error in predicting the labels given that the underlying data distribution $\D$ is known \cite{achille2018emergence,zhang2020dime}. For example, if $X$ does not provide any information for the label $Y$ such that 
$Y \independent X$, then the conditional entropy arrives its maximum: $H(Y|X)=H(Y)$, indicating the impossibility to guarantee a small prediction error. However, in the context of $\calH$-divergence \cite{ben2010theory}, this property \emph{can not be analyzed} since the label is determined by a \emph{fixed} labeling function, such that $H(Y|X) = \E_{x\sim\D(x)} H(h^{\star}(x)|X=x)\equiv 0$.

\textbf{Target Intrinsic Error: Upper Bound} In the context of DA, our goal is to ensure a small target risk, i.e. a small target intrinsic error is necessary. However, we never have the full target distribution $\calT(x,y)$, indicating the impossibility to directly estimate target intrinsic error $H(Y_t|X_t)$. In contrast, we can have the information of source distribution, as well as the relations of source and target distribution. Then we can derive the target intrinsic error is controlled by the label space size, as well as the source intrinsic error and Jensen-Shannon divergence of two distributions.   
This result is also consistent with our intuition and lower bound derived by Fano's inequality \cite{Polyanskiy2019}: a smaller label space $|\calY|$ is generally easier to learn, if the other conditions are identical.

\begin{theorem}\label{theorem: target_conditonal_entropy}
If $H(Y_s|X_s)\leq \epsilon$, the marginal and conditional distribution defined in Eq.~(\ref{eq:x_cond}) are close with $D_{\text{JS}}(\calS(x)\|\calT(x))\leq\delta_1$, and $\forall x$, $D_{\text{JS}}(\calS(y|X=x)\|\calT(y|X=x))\leq\delta_2$. Then the target intrinsic error can be upper bounded by:
\begin{equation*}
    H(Y_t|X_t)\leq \epsilon + \sqrt{\frac{\delta_2}{2}} + \frac{\sqrt{\delta_1}}{2} \log|\calY|.
\end{equation*}
\end{theorem}

\subsubsection{Application $\RN{2}$: Inherent Difficulty in Learning Open Set DA}\label{subsec:universal_da}
Our theory also proposes the analysis to understand when and what is difficult to transfer in Open Set DA, i.e. source and target domain share only a portion of label space \cite{cao2018partial,you2019universal,panareda2017open}, where $\calH$-divergence based theory completely fails to explain this scenario.

The key observation in the Open Set DA is that $\text{supp}\{\calT(y)\} \cap \text{supp}\{\calS(y)\}\neq\emptyset$. We suppose a small semantic conditional shift ($\forall y$, $D_{\text{JS}}(\calS(x|y)\|\calT(x|y))\leq\delta$), and a uniform label distributions over two different label spaces $\calY_1$ and $\calY_2$ such that $\calS(y)\sim\mathrm{Unif}(\calY_1)$, $\calT(y)\sim\mathrm{Unif}(\calY_2)$, $|\calY_1| = |\calY_2|=N$. We further assume the number of shared classes is $|\calY_1\cap\calY_2|=\alpha N$, $0<\alpha<1$. Then if the loss is binary and based on 
Theorem~\ref{theorem:lower_bound} and Eq.~(\ref{eq:y_cond}), the target risk can be bounded:
\begin{equation*}
   R_{\calS}(h) - \left(\sqrt{1-\alpha} + 2\sqrt{\delta}\right) \leq R_{\calT}(h) \leq  R_{\calS}(h) +  \frac{1}{\sqrt{2}} \left(\sqrt{1-\alpha} + 2\sqrt{\delta}\right).
\end{equation*}
When $\alpha\to 1$, $D_{\text{JS}}(\calT(y)\|\calS(y))\to 0$, the source risk is approaching the target risk from the two sides, then simply minimizing the source risk and further semantic conditional matching (see Sec.~\ref{sec:training}) can effectively control the target risk. By the contrary, if $\alpha\to 0$, the gap between target and source risk is large, indicating that a small source risk and semantic conditional shift no more guarantee a small target risk. From the practical perspective, less label overlapping means that it is harder to transfer the exact corresponding semantic conditional information from the source to the target.

\begin{table}[!t]
\caption{Empirical Methods for Bi-Directional Marginal/Conditional Shifts}
\centering
\label{tab:comp_conditional}
\begin{tabular}{@{}c|c|c|c|c@{}}
\toprule
\multicolumn{2}{l|}{Corollary 1}     & Source & Marginal Shift &  Conditional Shift      \\ \midrule
\multirow{2}{*}{Eq.(\ref{eq:x_cond})} & {Term} & $R_{\calS}(h)$ &  $D_{\text{JS}}(\calT(z)\|\calS(z))$  & $D_{\text{JS}}(\calT(y|z)\|\calS(y|z))$ \\ \cmidrule{2-5} 
                                      & Method & ERM  &  Feature Marginal Matching & {N/A} \\ \midrule
\multirow{2}{*}{Eq.(\ref{eq:y_cond})} & {Term} & {$R_{\calS}(h)$} & {$D_{\text{JS}}(\calT(y)\|\calS(y))$ } & $D_{\text{JS}}(\calT(z|y)\|\calS(z|y))$ \\ \cmidrule{2-5} 
                                      & Method &  \multicolumn{2}{|c|}{Label Marginal Shift Correction}  & Semantic Distribution Matching \\ 
\bottomrule
\end{tabular}
\end{table}

\section{Practical Principles for the Representation Learning}\label{sec:training}
In this section, we instantiate our theoretical framework with practical principles for designing DA algorithms in deep learning. Our results not only reaffirm the principles induced by $\calH$-divergence, but also motivate new DA practice in representation learning. 

We introduce the \emph{feature learning function} $g:\calX\to\mathcal{Z}$ and denote latent variable (feature) $z=g(x)$. Our objective is to find a representation function $g$ and classifier $h$, following the principles in Tab.~\ref{tab:comp_conditional}.
We also denote $\hat{\calS}(x,y)=\{(x_s^{i},y_s^{i})\}_{i=1}^{N_s}$, $\hat{\calT}(x)=\{x_t^{i}\}_{i=1}^{N_t}$ as the observed (empirical) distributions. 

\subsection{Inherent Practical Difficulty for Controlling Label Conditional Shift}\label{sec:algo_dann}
The upper bound in Eq. (\ref{eq:x_cond}) recovers the principles induced by $\calH$-divergence. Specifically, the domain adversarial training is equivalent to minimize the dual form of Jensen-Shannon divergence \cite{nowozin2016f}~i.e.~ $\min_{g} D_{\text{JS}}(\hat{\calT}(z)\|\hat{\calS}(z))$, the second principle in Eq.~(\ref{eq:x_cond}). 

However, domain adversarial training \emph{cannot} guarantee a small upper bound in Eq.~(\ref{eq:x_cond}). To this end, we can prove that merely minimizing $D_{\text{JS}}(\hat{\calT}(z)\|\hat{\calS}(z))$ can lead to an increase in the label conditional shift $D_{\text{JS}}(\hat{\calT}(y|z)\|\hat{\calS}(y|z))$ (\emph{see Appendix}), which can causes unexpected failures  \cite{wu2019domain}. 

Moreover, controlling the label condition shift is practically difficult. Because it requires two identical \emph{continuous} and high dimensional features such that $z_s = z_t$ with $z_s\in\hat{\calS}(z)$, $z_t\in\hat{\calT}(z)$, then minimizing $D_{\text{JS}}(\hat{\calT}(y|Z=z_s)\|\hat{\calS}(y|Z=z_t))$. In fact it is not trivial to find such feature pairs $z_s=z_t$, only from finite observational samples. 

\subsection{New Practical Principles}\label{sec:semantic}
According to Eq. (\ref{eq:y_cond}) in Corollary \ref{corollary_1}, the target risk can be alternatively bounded by $R_\calS(h)$, label marginal shift $D_{\text{JS}}(\calT(y)\|\calS(y))$, and semantic (feature) conditional shift $D_{\text{JS}}(\calT(z|y)\|\calS(z|y))$, which enables us to consider new principles in DA.

\paragraph{Semantic Conditional Distribution Matching} Different from controlling the label conditional shift $D_{\text{JS}}(
\hat{\calT}(y|Z=z)\|\hat{\calS}(y|Z=z))$, controlling the semantic (feature) conditional shift $D_{\text{JS}}(\hat{\calT}(z|Y=y)\|\hat{\calS}(z|Y=y))$ is practically more efficient, since \emph{labels are usually categorical variables with the finite classes}, comparing with continuous latent variable $Z$. However, there are no ground truth labels on the target domain, inducing the main issue in semantic conditional matching in DA. For addressing this concern, target \emph{pseudo labels} $Y_p$, estimated from the classifier, are introduced as the approximation of the real target label. 
Then following the insights of the third term in Eq.~(\ref{eq:y_cond}), the semantic conditional loss can be expressed as:
\begin{equation}
    \sum_{y} (\hat{\calS}(y) + \hat{\calT_p}(y))D_{\text{JS}}\left(\hat{\calT}(z|Y_p=y)\|\hat{\calS}(z|Y=y)\right),
    \label{eq:y_cond_upper}
\end{equation}
where $\hat{\calT_p}(y)$ is the target pseudo distribution  predicted by the neural network.  
We notice \cite{long2018conditional} alternatively encoded the label prediction information $h\circ g(x)$ as the conditional domain adversarial training, in order to implicitly minimize the conditional distribution divergence.  However, semantic conditional matching requires \emph{relative good pseudo-label prediction}. Otherwise the incorrect semantic (feature) feature alignment will lead to a negative transfer procedure for the target domain, during the learning phase.

\paragraph{Label Marginal Shift Correction} \emph{Is the semantic conditional matching sufficient to control the target risk?} From Eq.~(\ref{eq:y_cond}), the target risk is also controlled by label marginal shift. We can further extend this conclusion in the representation learning: if the semantic conditional distribution is matched, then the target risk is still controlled by the label marginal shift.
\begin{theorem}\label{theorem:label_shift_representation}
If any classifier $h$, feature learner $g$, and label $y\in\calY=\{-1,+1\}$ such that semantic conditional distribution is matched,  $D_{\text{JS}}(\calS(z|y),\calT(z|y))=0$, then the target risk can be bounded:
\begin{equation*}
      R_{\calS}(h \circ g)- \sqrt{2 D_{\text{JS}}(\calS(y),\calT(y))}  \leq R_{\calT}(h \circ g) \leq R_{\calS}(h\circ g) + \sqrt{2 D_{\text{JS}}(\calS(y),\calT(y))},
\end{equation*}
where $R_{\calS}(h \circ g) = R_{\calS}(h(g(x),y))$ is the expected risk over the classifier $h$ and feature learner $g$.
\end{theorem}
As Theorem \ref{theorem:label_shift_representation} suggests, we need to control label marginal shift $D_{\text{JS}}(\calT(y)\|\calS(y))$. Therefore we adopt the popular label re-weighted loss \cite{cortes2010learning}: $\hat{R}^{\alpha}_{\calS}(h \circ g) = \sum_{(x_s,y_s)\sim\hat{\calS}(x,y)} \alpha(y_s) L(h(g(x_s),y_s))$ with $\alpha(y) = \frac{\calT(y)}{\calS(y)}$. In addition, we can further prove the empirical re-weighted loss converges to $R_{\calT}(h\circ g)$, if $D_{\text{JS}}(\calS(z|y),\calT(z|y))=0$~(see Appendix). 
As for estimation the label weight $\hat{\alpha}$ from the data, several approaches have been proposed, e.g.\ Black Box Shift Learning (BBSL) \cite{lipton2018detecting} or Regularized Learning under Label Shift (RLSS) \cite{azizzadenesheli2018regularized}.

\paragraph{Feature Marginal Matching as the Constraint}
Although the aforementioned principles are theoretically appealing, but we practically use the pseudo-label $Y_p$ for the semantic conditional matching $D_{\text{JS}}(\hat{\calT}(z|Y_p=y)\|\hat{\calS}(z|Y=y))$, 
which can lead to the negative transfer in the training loop
if we face the poor pseudo-label prediction.

Can we derive the principle to recognize the poor pseudo-label prediction during the learning? Theorem~\ref{theorem: small_marginal} reveals one consequence of the poor target pseudo-label prediction: it can lead to a large empirical feature marginal divergence $D_{\text{JS}}(\hat{\calS}(z)\|\hat{\calT}(z))$ (in Eq.~(\ref{eq:x_cond})), under mild conditions.

\begin{theorem}\label{theorem: small_marginal}
We denote $\hat{\calS}_p(y),\hat{\calT}_p(y)$ are the prediction output (pseudo-label) distributions.
If we have such a ``bad'' pseudo-label prediction such that $D_{\text{JS}}(\hat{\calT}(y)\|\hat{\calT}_{p}(y)) = P$, small source prediction error $D_{\text{JS}}(\hat{\calS}(y)\|\hat{\calS}_{p}(y))\leq \epsilon_1$ and small label ground truth empirical distribution divergence $D_{\text{JS}}(\hat{\calS}(y)\|\hat{\calT}(y))\leq \epsilon_2$, then the feature marginal divergence on the latent space $Z$ can be lower bounded by: 
\begin{equation*}
    D_{\text{JS}}(\hat{\calS}(z)\|\hat{\calT}(z))\geq (\sqrt{P}-\sqrt{\epsilon_1}-\sqrt{\epsilon_2})^2.
\end{equation*}
\end{theorem}
From Theorem \ref{theorem: small_marginal}, if $P\to 1$ and $\epsilon_1,\epsilon_2$ are small, $D_{\text{JS}}(\hat{\calS}(z)\|\hat{\calT}(z))$ can be very large. Therefore we add the constraint $\D_{\text{JS}}(\hat{\calS}(z)\|\hat{\calT}(z))\leq \kappa$ as a broad adaptation step, to prevent the poor pseudo-label prediction (a.k.a. a large $P$).

\paragraph{Practical Guideline} Based on these three-principles, we propose a generic and iterative practical framework, where parameter optimization and pseudo-label prediction steps are conducted iteratively. 

Moreover, we would like to emphasize the realization of each principle is flexible. For example, the distribution matching can be done through either adversarial training by introducing the auxiliary domain discriminator $d$ or parametric distribution matching (e.g. statistical moment matching approach). More empirical choices can be found in the Appendix.

\begin{tcolorbox}[grow to left by=0.3cm,grow to right by=0.2cm]
\textbf{Parameter Optimization Step} (fixed Pseudo-Labels) classifier $h$ and feature extractor $g$:
\begin{equation*}
    \begin{split}
         & \min_{h,g}\quad \underbrace{\hat{R}^{\hat{\alpha}}_{\calS}(h(g(x),y))}_{(\RN{1})} + \underbrace{\sum\nolimits_{y} (\hat{\calS}(y) + \hat{\calT}_p(y))D_{\text{JS}}\left(\hat{\calT}(g(x)|Y_p=y)\|\hat{\calS}(g(x)|Y=y)\right)}_{(\RN{2})} \\
         & \text{s.t.}\quad\underbrace{D_{\text{JS}}(\hat{\calT}(g(x))\|\hat{\calS}(g(x)))\leq \kappa}_{(\RN{3})}
    \end{split}
\end{equation*}
$(\RN{1})$~Label marginal shift correction:~$\hat{R}^{\hat{\alpha}}_{\calS}(h(g(x),y)) = \sum_{(x_s,y_s)\sim\hat{\calS}} \hat{\alpha}(y_s) L(h(g(x_s),y_s))$; $(\RN{2})$~Semantic conditional matching, aligning the semantic feature; $(\RN{3})$~feature marginal matching as the constraint, a broad adaptation step to prevent a poor initialization of pseudo-label prediction.
\paragraph{Pseudo-Label Prediction Step} (fixed Parameters):   $y^{p}$, $\hat{\alpha}$, $\hat{\calT}_p(y)$ \\
$y^{p}$,$\hat{\calT}_p(y)$ are pseudo-labels and distributions on the target domain. $\hat{\alpha}$ is the reweighting coefficient. 
\end{tcolorbox}

\section{Related Work}
An important aspect in DA is to establish the proper distribution discrepancy. Existing works have proposed hypothesis based metrics such as $\calH$-divergence \cite{ben2010theory}, Distribution discrepancy \cite{cortes2019adaptation}, Margin disparity discrepancy \cite{zhang2019bridging}, as well as the statistical divergence such as R\'enyi divergence \cite{mansour2009multiple,germain2016new,hoffman2018algorithms}, Wasserstein distance \cite{redko2017theoretical}. However, these theoretical results mainly focus on the feature marginal discrepancy, which are generally difficult to handle the semantic conditional shift.

Another key factor is to understand and handle the conditional shift. From a series of theoretical results \cite{ben2010theory,ben2010impossibility,ben2014domain,germain2013pac,johansson2019support}, a small joint optimal risk $\beta$ is crucial to ensure a small target risk. Following this line, \cite{zhao2019learning} adopted Jensen-Shannon divergence to derive the lower bound of $\beta$, indicating the necessarily of considering conditional shift. However, it is still not clear how the algorithms explicitly guarantee a small $\beta$. Indeed, our work further extend this by proving the theoretical results without $\beta$, which enable the possible practice to explicitly control the target risk. 
\cite{zhang2013domain,gong2016domain} analyzed feature conditional shift from the causal prospective, which is generally difficult to adapt in the large-scale dataset. \cite{li2019target,tan2019generalized,long2013transfer,saito2017asymmetric,pmlr-v89-redko19a,chen2019progressive,xie2018learning,cai2019learning} proposed empirical strategies for eliminating conditional shift, which speculated one or two principles to improve the empirical performance. We formally demonstrate the unified three-principles, as the way to control the target risk.     

\section{Experimental Validations}\label{sec:empirical_results}
We validate the proposed guideline by employing the \emph{existing methods} for each principle. We aim to show \emph{whether applying the unified three principles is better than merely considering only one or two of them.} We delegate algorithm details, dataset descriptions experimental settings in the Appendix.

\textbf{Algorithm Selection}\quad$(\RN{1})$~Source Re-weighting loss. We estimate the $\hat{\alpha}$ by BBSL approach \cite{lipton2018detecting} with $\hat{\alpha} = \hat{C}^{-1}\hat{\calT}_p$, where $\hat{C}$ is source prediction confusion matrix with $\hat{C}[i,j] = \Proba(h(g(x_s))=i, y_s=j)$. $(\RN{2})$ Semantic Conditional Matching. For each label $Y=y$, we align their first order statistics (feature centroid): $\|\frac{1}{|\# y_s=y|}\sum_{(x_s,y_s)} \delta_{\{y_s=y\}} g(x_s) - \frac{1}{|\# y^{p}_t=y|}\sum_{(x_t,y_t^{p})} \delta_{\{y_t^{p}=y\}} g(x_t)\|_2$, which is the approximation of $d_{\text{TV}}$, the upper bound of $D_{\text{JS}}$. $(\RN{3})$~feature marginal matching as the constraint. We simply adopt the Lagrangian relaxation for treating the constraint as the regularization, with $\kappa$ the hyperparameter. We use adversarial loss in Eq.~(\ref{eq: gans_loss}) to estimate $D_{\text{JS}}(\hat{\calT}(z)\|\hat{\calS}(z))$. 

\textbf{Brief Experimental Settings}\quad We evaluate the framework on Office-31 \cite{saenko2010adapting} and CLEF \cite{villegas2015general}, the complex and real image datasets. We implement the framework based on the Pre-trained AlexNet \cite{krizhevsky2012imagenet} and compare the baselines of merely considering marginal (DANN\cite{ganin2016domain}), conditional (CDAN\cite{long2018conditional}), and parts of our principles. We repeat the experiments five times and report the average and variance.

\textbf{Results and Analysis}\quad We report the empirical performances in Tab. \ref{tab:office_31} and \ref{tab:clef_dataset}.
\begin{table}[ht]
\caption{Accuracy ($\%$) on Office-31 Dataset}
    \centering
    \resizebox{0.9\textwidth}{!}{
    \begin{tabular}{l|cccccc|cc}  
\toprule
Method  & A $\rightarrow$ D  & A $\rightarrow$ W & D $\rightarrow$ W  & W $\rightarrow$ D & W $\rightarrow$ A & D $\rightarrow$ A &  Ave \\
\midrule
Without DA  &  63.8$\pm$0.5   &  61.6$\pm$0.5& 95.4$\pm$0.3 & 99.0$\pm$0.2 & 49.8$\pm$0.4  & 51.1$\pm$0.6  & 70.1   \\
DANN~\cite{ganin2016domain} & 72.3$\pm$0.3   &  73.0$\pm$0.5& 96.4$\pm$0.3 & 99.2$\pm$0.3 &  51.2$\pm$0.5 & 52.4$\pm$0.4  & 74.1   \\ 
CDAN~\cite{long2018conditional} & 76.3$\pm$0.1 & 78.3$\pm$0.2  & 97.2$\pm$0.1 & \textbf{100.0}$\pm$0.0 & 57.5$\pm$0.4 &  57.3$\pm$0.2  & 77.7    \\ \hline
$(\RN{1}+\RN{3})$  & 72.6$\pm$0.4  & 73.5$\pm$0.4 &  96.2$\pm$0.2  & 99.3$\pm$0.5  & 51.4$\pm$0.2  & 52.8$\pm$0.5  & 74.3  \\
$(\RN{1}+\RN{2})$  &  75.3$\pm$0.7   & 79.4$\pm$1.1 & 97.1$\pm$0.5 & 97.5$\pm$0.5 & 58.2$\pm$0.9  & 61.8$\pm$0.8  & 78.2  \\
$(\RN{2}+\RN{3})$  & 75.7$\pm$0.1  & 79.2$\pm$0.7& 96.8$\pm$0.1 & 99.8$\pm$0.1 & 59.5$\pm$0.4  & 58.7$\pm$0.3  & 78.3  \\
\hline
$(\RN{1}+\RN{2}+\RN{3})$ & \textbf{76.7}$\pm$0.4 & \textbf{80.8}$\pm$0.4 & \textbf{97.5}$\pm$0.2 & 99.8$\pm$0.1 & \textbf{59.8}$\pm$0.4 & \textbf{62.3}$\pm$0.2 & \textbf{79.5} \\
\bottomrule
\end{tabular}}
\label{tab:office_31}
\end{table}

\begin{table}[ht]
\caption{Accuracy ($\%$) on CLEF Dataset}
    \centering
    \resizebox{0.9\textwidth}{!}{
    \begin{tabular}{l|cccccc|cc}  
\toprule
Method  & I $\to$ C  & I $\rightarrow$ P & C $\rightarrow$ I  & P $\rightarrow$ I & C $\rightarrow$ P & P $\rightarrow$ C &  Ave \\
\midrule
Without DA  & 84.3$\pm$0.2  & 66.2$\pm$0.2 & 71.3$\pm$0.4 & 70.0$\pm$0.2 & 59.3$\pm$0.5 & 84.5$\pm$0.3   & 73.9   \\
DANN~\cite{ganin2016domain} & 89.0$\pm$0.4  & 66.5$\pm$0.3 & 79.8$\pm$0.4 & 81.8$\pm$0.3 & 63.5$\pm$0.5  &  88.7$\pm$0.3  & 78.2   \\
CDAN~\cite{long2018conditional} & 91.8$\pm$0.2  & 67.7$\pm$0.3 & 81.5$\pm$0.2 & 83.3$\pm$0.1 & 63.0$\pm$0.2  & 91.5$\pm$0.3   & 79.8  \\\hline
$(\RN{1}+\RN{3})$  & 89.3$\pm$0.2  & 67.0$\pm$0.6 &  80.0$\pm$0.7  & 81.9$\pm$0.3  & 62.9$\pm$0.4  & 89.2$\pm$0.2  & 78.4  \\
$(\RN{1}+\RN{2})$  & 90.2$\pm$0.5  & 66.7$\pm$0.6 & 80.3$\pm$0.5 & 82.7$\pm$0.7 & 62.5$\pm$0.7  & 90.7$\pm$0.6  & 78.8  \\
$(\RN{2}+\RN{3})$   & 91.5$\pm$0.1   & 67.3$\pm$0.3 & 81.7$\pm$0.3 & 82.8$\pm$0.2 & 63.5$\pm$0.4 & 91.2$\pm$0.2   & 79.9   \\\hline
$(\RN{1}+\RN{2}+\RN{3})$ & \textbf{92.1}$\pm$0.2 & \textbf{68.2}$\pm$0.2 & \textbf{82.1}$\pm$0.2 & \textbf{84.0}$\pm$0.2 & \textbf{64.2}$\pm$0.2 & \textbf{91.9}$\pm$0.1 &  \textbf{80.4} \\
\bottomrule
\end{tabular}}
\label{tab:clef_dataset}
\end{table}

The empirical results indicate the improved performance on the unified principles, comparing with merely one or two principles. 
We observe the empirical benefit of semantic conditional matching $(\RN{2})$ is relative more notable. Besides, the principles without feature marginal matching $(\RN{3})$ not only lead to a drop in the performance, but also increase its instability with a relative higher variance.

\begin{figure}
  \centering
  \begin{subfigure}{0.4\textwidth}
  \centering
     \includegraphics[scale=0.4]{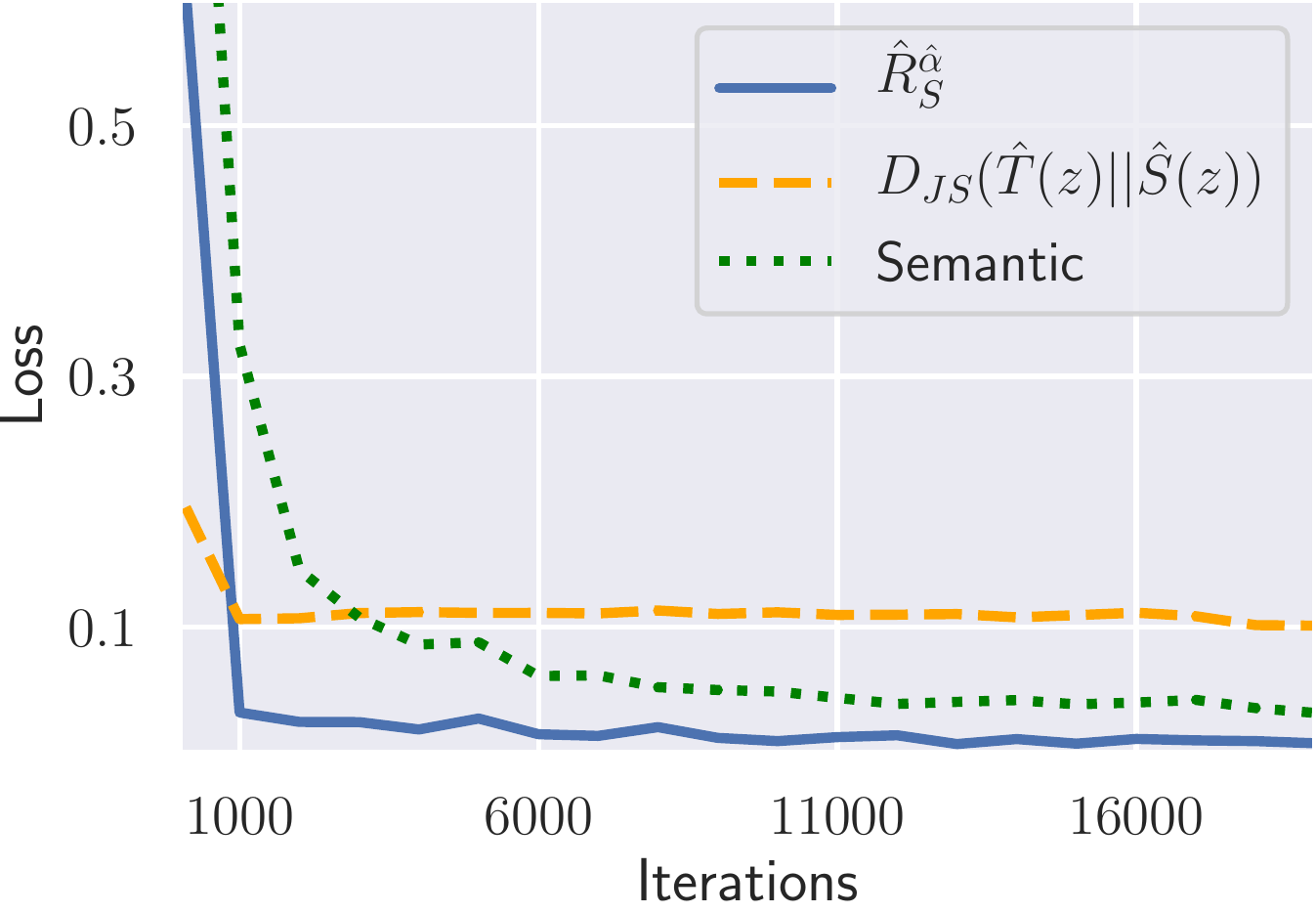}
     \caption{Evolution of Each Principle}
  \end{subfigure}
  \begin{subfigure}{0.4\textwidth}
  \centering
     \includegraphics[scale=0.4]{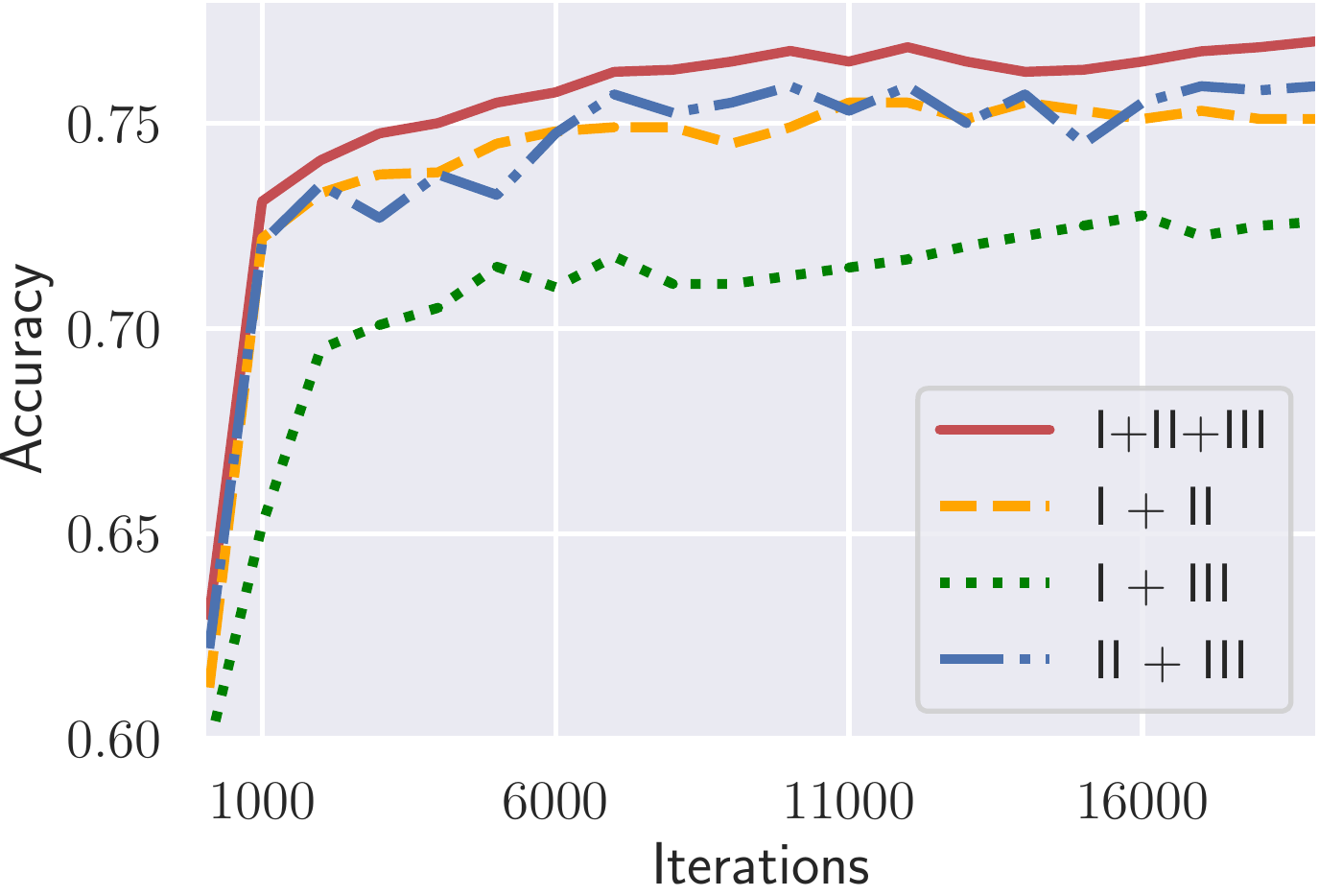}
     \caption{Ablation Study}
  \end{subfigure}
  \caption{Properties of Proposed Principles (Task A $\rightarrow$ D)}
  \label{fig:analysis}
\end{figure}

Fig.~\ref{fig:analysis} further reveals the properties of the proposed  principles. Specifically, Fig.~\ref{fig:analysis}(a) shows the evolution of each principle (loss) during the training, which is exact coherent with the goals in the guideline. The semantic conditional shift (Principle $\RN{2}$) and the weighted source classification error (Principle $\RN{1}$) gradually diminish and $D_{\text{JS}}(\hat{\calT}(z)\|\hat{\calS}(z))$ (Principle $\RN{3}$) restricts within a small value. In addition, we trace the target domain prediction accuracy of different principles combinations in Fig. \ref{fig:analysis}(b), for demonstrating the impact of each principle. The results indicate the importance of considering semantic (feature) conditional distribution matching $(\RN{2})$, with a significant performance influence $(\sim 4.2\%)$. On the other hand, the influences of principle $(\RN{1})$ and $(\RN{3})$ are relatively modest $(\sim 1.3\%)$. We think the reason is the employed pre-trained network for the DA training strategies in the complex datasets, which already has a relative good initial prediction. 

\section{Conclusion}
We proposed a new theoretical framework based on Jensen-Shannon divergence for analyzing DA problems. Our theory established bi-directional marginal/conditional shifts for the target risk bound. We further demonstrated its flexibility in various theoretical and algorithmic applications. It is worth mentioning that our theoretical framework is not only suitable for DA, but also extendable to analyzing the real shift problems such as fair representation learning~\cite{louizos2015variational,edwards2015censoring}, individual treatment effect estimation~\cite{shalit2017estimating}. We anticipate that our theory can open up a pathway towards new algorithm designs for DA, driven by the advantages of fundamental understanding. 

\bibliography{ref}
\bibliographystyle{unsrt}

\appendix
\section{$\calH$-Divergence v.s. Jensen-Shannon Divergence}
\subsection{Counterexample One}
We take the example proposed by \cite{ben2010impossibility} (Example 6), which has already computed the $d_{\calH}(\calS(x),\calT(x)) = \xi$. However, since $\text{supp}(\calS(x))\cap\text{supp}(\calT(x))=\emptyset$, $D_{\text{JS}}(\calS(x)\|\calT(x))=1$.
\subsection{Counterexample Two}
We have $\calS = \mathrm{Unif}\{1,2,3\}$ and $\calT = \{ \Proba(X=1)=\frac{1}{4}, \Proba(X=2)= \frac{1}{2}, \Proba(X=3)=\frac{1}{4}\}$. 
\paragraph{Computing $d_{\calH}$} It is also related to the optimal classification error.
\begin{equation*}
\mathrm{err}(h) = 
     \begin{cases}
       1/2   &\quad\text{if}\quad t<1, t>3   \\
       11/24 &\quad\text{if}\quad 1<t<2 \\
       13/24   &\quad\text{if}\quad 2<t<3 \\
     \end{cases}
\end{equation*}
Then the $\calH$ divergence is $d_{\calH}(\calT(x),\calS(x)) = 1-2\min_{h}[\mathrm{err}(h)] = \frac{1}{12}\approx 0.0833$
\paragraph{Computing $D_{\text{JS}}(\calT(x)\|\calS(x))$} 
Since the two distributions hold the same support, we can compute the mixture distribution $\calM = \{ \Proba(X=1)=\frac{7}{24}, \Proba(X=2)= \frac{5}{12}, \Proba(X=3)=\frac{7}{24}\}$, 
We can compute the Jensen-Shannon divergence:
\begin{equation*}
    \begin{array}{cc}
     & D(\calS\|\calM) = \frac{1}{3}\log(\frac{1/3}{7/24}) +  \frac{1}{3}\log(\frac{1/3}{5/12}) + \frac{1}{3}\log(\frac{1/3}{7/24}) \approx 0.02110 \\
     & D(\calT\|\calM) = \frac{1}{4}\log(\frac{1/4}{7/24}) +  \frac{1}{2}\log(\frac{1/2}{5/12}) + \frac{1}{4}\log(\frac{1/4}{7/24})\approx 0.02032
\end{array}
\end{equation*}
Then $D_{\text{JS}}(\calT(x)\|\calS(x)) = \frac{1}{2}(0.0211+0.02032) = 0.0207$.
In this scenario, the  $D_{\text{JS}}(\calT(x)\|\calS(x)) < d_{\calH}(\calT(x),\calS(x))$, therefore, the $D_{\text{JS}}$ can not be viewed as an upper bound of $d_{\calH}$.

\subsection{Discussions}
We notice \cite{ben2010theory} analogously proposed the $d_{\calH\Delta\calH}$ divergence to measure domain discrepancy. However, as it pointed out (in Sec.~7.2 of \cite{ben2010theory}), it is also impossible to exactly estimate this discrepancy. As the consequence, \cite{ben2010theory} still adopted $d_{\calA}$ distance to approximate, recovering the same empirical strategy as $\calH$ divergence.   

\begin{table}[h]
\centering
\caption{Different DA Theories.  Data Generation: distribution assumption; General Loss: adaptability on \emph{non binary loss}; Stochastic Setting: adaptability on stochastic conditional generation procedure.}
\label{tab:comp_theorem}
\begin{tabular}{@{}c|c|c|c@{}}
\toprule
 Divergence & Data Generation &  General Loss  &  Stochastic Settings    \\ \midrule
 $\calH$-divergence \cite{ben2010theory} & \multirow{2}{*}{\makecell{$x\sim\D(x)$, $y= h^{\star}(x)$\\ $|\calY|=2$}} &      $\times$  & $\times$    \\ \cmidrule(r){1-1} \cmidrule(lr){3-3}\cmidrule(lr){4-4}
 Discrepancy \cite{mansour2009domain} &                 & $\surd$ &     $\times$              \\ \midrule
Jensen-Shannon & \makecell{$x\sim\D(x)$, $y\sim \D(y|x)$\\ $|\calY|\geq 2$}                  & $\surd$               &          $\surd$      \\ \bottomrule
\end{tabular}
\end{table}

\section{Domain Adaptation: Upper Bound}
We first prove an intermediate lemma:
\begin{tcolorbox}[grow to left by=0.3cm,grow to right by=0.3cm]
\begin{lemma}
Let $Z\in\mathcal{Z}$ be the real valued integrable random variable, let $P$ and $Q$ are two distributions on a common space $\mathcal{Z}$ such that $Q$ is absolutely continuous w.r.t. $P$. If for any function $f$ and $\lambda\in\R$ such that $\E_P[e^{\lambda(f(z)-\E_P(f(z))}]<\infty$, then we have:
\begin{equation*}
   \lambda (\E_{Q} f(z) - \E_{P}f(z)) \leq D_{\text{KL}}(Q\|P) + \log \E_P[e^{\lambda(f(z)-\E_P(f(z))}]
\end{equation*}
Where $D_{\text{KL}}(Q\|P)$ is the Kullback–Leibler divergence between distribution $Q$ and $P$, and the equality arrives when $f(z)= \E_{P} f(z) + \frac{1}{\lambda}\log(\frac{d Q}{d P})$.
\end{lemma}
\end{tcolorbox}

\begin{proof}
We let $g$ be \textbf{any} function such that $\E_P[e^{g(z)}]<\infty$, then we define a random variable $Z_g(z) = \frac{e^{g(z)}}{\E_P[e^{g(z)}]}$, then we can verify that $\E_{P}(Z_g) =1$. We assume another distribution $Q$ such that $Q$ (with distribution density $q(z)$) is absolutely continuous w.r.t. $P$ (with distribution density $p(z)$), then we have:
\begin{equation*}
\begin{split}
     \E_{Q}[\log Z_g] & = \E_{Q}[\log\frac{q(z)}{p(z)} + \log(Z_g\frac{p(z)}{q(z)})] \\
     & = D_{\text{KL}}(Q\|P) + \E_{Q}[\log(Z_g\frac{p(z)}{q(z)})]\\
     & \leq D_{\text{KL}}(Q\|P) + \log\E_{Q}[\frac{p(z)}{q(z)}Z_g]\\
     & = D_{\text{KL}}(Q\|P) + \log \E_{P}[Z_g]
\end{split}
\end{equation*}
Since $\E_{P}[Z_g] = 1$ and according to the definition we have $\E_{Q}[\log Z_g] = \E_{Q}[g(z)] - \E_{Q}\log\E_{P}[e^{g(z)}] = \E_{Q}[g(z)] - \log\E_{P}[e^{g(z)}]$ (since $\E_{P}[e^{g(z)}]$ is a constant w.r.t. $Q$) and we therefore have:

\begin{equation}
    \E_{Q}[g(z)] \leq  \log\E_{P}[e^{g(z)}] + D_{\text{KL}}(Q\|P)
    \label{change_of_measure}
\end{equation}
Since this inequality holds for any function $g$ with finite moment generation function, then we let $g(z) = \lambda( f(z)-\E_P f(z))$ such that $\E_P[e^{f(z)-\E_P f(z)}]<\infty$. Therefore we have $\forall \lambda$ and $f$ we have:
\begin{equation*}
    \E_{Q}\lambda(f(z)-\E_P f(z)) \leq D_{\text{KL}}(Q\|P) + \log \E_P[e^{\lambda(f(z)-\E_P f(z)}]
\end{equation*}
Since we have $\E_{Q}\lambda(f(z)-\E_P  f(z)) = \lambda \E_{Q} (f(z)-\E_P f(z))) = \lambda (\E_{Q} f(z) - \E_{P} f(z))$, therefore we have:
\begin{equation*}
   \lambda (\E_{Q} f(z) - \E_{P} f(z)) \leq D_{\text{KL}}(Q\|P) + \log \E_P[e^{\lambda (\E_{Q} f(z) - \E_{P} f(z))}]
\end{equation*}
As for the attainment in the equality of Eq.(\ref{change_of_measure}), we can simply set  $g(z) = \log(\frac{q(z)}{p(z)})$, then we can compute $\E_{P}[e^{g(z)}]=1$ and the equality arrives. Therefore in Lemma 1, the equality reaches when $\lambda(f(z)- \E_{P} f(z)) = \log(\frac{d Q}{d P})$.
\end{proof}

In the classification problem, we define the observation pair $z=(x,y)$. We also define the loss function $\ell(z)=L\circ h(z)$ with deterministic hypothesis $h$ and prediction loss function $L$. Then for abuse of notation, we simply denote the loss function $\ell(z)$ in this part. 

\begin{tcolorbox}[grow to left by=0.3cm,grow to right by=0.3cm]
Supposing the prediction loss $L$ is bounded with interval $G$ with $G = \max(L) -\min(L)$, then the expected risk in the target domain can be upper bounded by:
\begin{equation*}
    R_{\calT}(h) \leq R_{\calS}(h) + \frac{G}{\sqrt{2}}\sqrt{D_{\mathrm{JS}}(\calT\|\calS)}
\end{equation*}
Where $D_{\mathrm{JS}} = \frac{1}{2}\big(D(\calT\|\frac{1}{2}(\calT+\calS)) +D(\calS\|\frac{1}{2}(\calT+\calS))\big) $ is the joint Jensen-Shannon divergence. 
\end{tcolorbox}

\begin{proof}
According to Lemma 1, $\forall \lambda>0$ we have: 
\begin{equation}
    \E_Q f(z) - \E_P f(z) \leq \frac{1}{\lambda} (\log\E_{P}~e^{[\lambda(f(z)-\E_{P}f(z))]} + D_{\text{KL}}(Q\|P))
    \label{sub_1}
\end{equation}

\noindent And $\forall \lambda<0$ we have:
\begin{equation}
    \E_Q f(z) - \E_P f(z) \geq \frac{1}{\lambda} (\log\E_{P}~e^{[\lambda(f(z)-\E_{P}f(z))]} + D_{\text{KL}}(Q\|P))
    \label{sub_2}
\end{equation}

Then we introduce an intermediate distribution $\calM(z) = \frac{1}{2}(\calS(z) + \calT(z))$, then $\text{supp}(\calS)\subseteq\text{supp}(\calM)$ and $\text{supp}(\calT)\subseteq\text{supp}(\calM)$, and let $f=\ell$. Since the random variable $\ell$ is bounded through $G =  \max(L) -\min(L)$, then according to \cite{wainwright2019high}(Chapter 2.1.2), $\ell-\E_{P}\ell$ is sub-Gaussian with parameter at most $\sigma = \frac{G}{2}$, then we can apply Sub-Gaussian property to bound the $\log$ moment generation function:
\begin{equation*}
    \log\E_{P}~e^{[\lambda(\ell(z)-\E_{P}\ell(z))]} \leq \log e^{\frac{\lambda^2\sigma^2}{2}} \leq \frac{\lambda^2G^2}{8}.
\end{equation*}

\noindent In Eq.(\ref{sub_1}), we let $Q = \calT$ and $P=\calM$, then $\forall \lambda>0$ we have:
\begin{equation}
    \E_{\calT}~\ell(z) - \E_{\calM}~\ell(z) \leq \frac{G^2\lambda}{8} +  \frac{1}{\lambda}D_{\text{KL}}(\calT\|\calM)
    \label{sub_3}
\end{equation}

\noindent In Eq.(\ref{sub_2}), we let $Q = \calS$ and $P=\calM$, then $\forall \lambda<0$ we have:
\begin{equation}
    \E_{\calS}~\ell(z) - \E_{\calM}~\ell(z) \geq \frac{G^2\lambda}{8} +  \frac{1}{\lambda}D_{\text{KL}}(\calS\|\calM)
    \label{sub_4}
\end{equation}

\noindent In Eq.(\ref{sub_3}), we denote $\lambda=\lambda_0>0$ and $\lambda=-\lambda_0<0$ in Eq.(\ref{sub_4}).
Then Eq.(\ref{sub_3}),  Eq.(\ref{sub_4}) can be reformulated as:
\begin{equation}
\begin{split}
    & \E_{\calT}~\ell(z) - \E_{\calM}~\ell(z) \leq \frac{G^2\lambda_0}{8} +  \frac{1}{\lambda_0}D_{\text{KL}}(\calT\|\calM)\\
    & \E_{\calM}~\ell(z) - \E_{\calS}~\ell(z) \leq \frac{G^2\lambda_0}{8} +  \frac{1}{\lambda_0}D_{\text{KL}}(\calS\|\calM)
\end{split}
    \label{sub_5}
\end{equation}
Adding the two inequalities in Eq.(\ref{sub_5}), we therefore have:
\begin{equation}
    \E_{\calT}~\ell(z)  \leq \E_{\calS}~\ell(z) + \frac{1}{\lambda_0} \big(D_{\text{KL}}(\calS\|\calM) + D_{\text{KL}}(\calT\|\calM) \big) + \frac{\lambda_0}{4}G^2 
\end{equation}
Since the inequality holds for $\forall \lambda_0$, then by taking $\lambda_0 = \frac{2}{G}\sqrt{D_{\text{KL}}(\calS\|\calM) + D_{\text{KL}}(\calT\|\calM)}$ we finally have:
\begin{equation}
      \E_{\calT}~\ell(z)  \leq \E_{\calS}~\ell(z) + \frac{G}{\sqrt{2}}\sqrt{D_{\text{JS}}(\calT\|\calS)}
    \label{sub_6}
\end{equation}
\end{proof}

\subsection{Extension to Unbounded Loss}
The advantage of proposed theory can be naturally extended to the unbounded loss. 

\begin{tcolorbox}[grow to left by=0.3cm,grow to right by=0.3cm]
\begin{corollary}[Sub-Gaussian Upper Bound]
If the loss function satisfies $\sigma$-Sub Gaussian property: $\log\E_{P}~e^{[\lambda(\ell(z)-\E_{P}\ell(z))]} \leq \frac{\lambda^2\sigma^2}{2}$, then the expected risk in the target domain can be upper bounded by:
\begin{equation*}
    R_{\calT}(h) \leq R_{\calS}(h) + \sigma\sqrt{2D_{\mathrm{JS}}(\calT\|\calS)}
\end{equation*}
\end{corollary}
\end{tcolorbox}

\begin{proof}
The proof is trivial by simply plugging in the Sub-Gaussian condition in the moment generation function. 
\end{proof}

\begin{tcolorbox}[grow to left by=0.3cm,grow to right by=0.3cm]
\begin{corollary}[Sub-Gamma Upper Bound] If the loss function satisfies $(\sigma,a)$-Sub Gamma property: $\log\E_{P}~e^{[\lambda(\ell(z)-\E_{P}\ell(z))]} \leq \frac{\lambda^2 \sigma}{2(1-a|\lambda|)}$, for $0<|\lambda|<\frac{1}{a}$. Then the expected risk in the target domain can be upper bounded by:
\begin{equation*}
     R_{\calT}(h) \leq  R_{\calS}(h) + (\sigma+1)\sqrt{2D_{\text{JS}}(\calT\|\calS)} + 2a D_{\text{JS}}(\calT\|\calS)
\end{equation*}
\end{corollary}
\end{tcolorbox}

\begin{proof}
For the same step for the moment generation function, by taking $\lambda_0\in(0,\frac{1}{a})$, then analogously we have:
\begin{equation*}
\begin{split}
    & \E_{\calT}~\ell(z) - \E_{\calM}~\ell(z) \leq \frac{\lambda_0\sigma}{2(1-a\lambda_0)} +  \frac{1}{\lambda_0}D_{\text{KL}}(\calT\|\calM)\\
    & \E_{\calM}~\ell(z) - \E_{\calS}~\ell(z) \leq  \frac{\lambda_0\sigma}{2(1-a\lambda_0)} +  \frac{1}{\lambda_0}D_{\text{KL}}(\calS\|\calM)
\end{split}
\end{equation*}
Therefore we have 
\begin{equation*}
\begin{split}
    \E_{\calT}~\ell(z) - \E_{\calS}~\ell(z) & \leq \frac{\lambda_0\sigma}{(1-a\lambda_0)} +  \frac{1}{\lambda_0}\left(D_{\text{KL}}(\calT\|\calM)+ D_{\text{KL}}(\calS\|\calM)\right)\\
    & = \frac{\lambda_0\sigma}{(1-a\lambda_0)} +  \frac{1}{\lambda_0}\left(2D_{\text{JS}}(\calT\|\calS)\right)
\end{split}
\end{equation*}
We let $\lambda_0 = \frac{\sqrt{2D_{\text{JS}}(\calT\|\calS)}}{\sigma+a\sqrt{2D_{\text{JS}}(\calT\|\calS)}}\in (0,\frac{1}{a})$ and we can simplify the upper bound as:
\begin{equation*}
     \E_{\calT}~\ell(z) - \E_{\calS}~\ell(z) \leq (\sigma+1)\sqrt{2D_{\text{JS}}(\calT\|\calS)} + 2a D_{\text{JS}}(\calT\|\calS)
\end{equation*}
\end{proof}

The extended upper bounds can be much tighter than the conclusion in Theorem 1, particularly when the loss is in a large range with a small variance.

\section{Domain Adaptation Theory: Lower Bound}
We firstly introduce several information theoretical tools:
\begin{lemma}[Pinsker's inequality]If $P$ and $Q$ are two probability distribution on the measurable space $(\Omega,\mathcal{F})$, then 
\begin{equation*}
    \mathrm{TV}(P,Q) \leq \sqrt{2D_{\text{KL}}(P\|Q)}
\end{equation*}
Where $D(P\|Q)_{\text{KL}}$ is the Kullback–Leibler divergence between distribution $P$ and $Q$ and $TV(P\|Q) = \sum_{z}|P(z)-Q(z)|$
\end{lemma}

\begin{lemma}\cite{Polyanskiy2019}[$f$-divergence data processing inequality]
Consider a channel that produces $Y$ given $X$ on the deterministic function $g$. If $P_Y$ is the distribution of $Y$ when $X$ is generated by $P_X$ and $Q_Y$ is the distribution of $Y$ when $X$ is generated by $Q_X$, then for any $f$-divergence $D_{f}(\cdot\|\cdot)$:
\begin{equation*}
    D_{f}(P_Y\|Q_Y) \leq D_{f}(P_{X}\|Q_{X})
\end{equation*}
\end{lemma}

\begin{tcolorbox}[grow to left by=0.3cm,grow to right by=0.3cm]
If we restrict the zero-one loss $L\in \{0,1\}$, then we can prove the target risk be lower bounded by:
\begin{equation*}
    R_{\calT}(h) \geq R_{\calS}(h) - \sqrt{D_{\mathrm{JS}}(\calT\|\calS)}
\end{equation*}
\end{tcolorbox}

\begin{proof}
Again we denote the observation pair $z=(x,y)$. For abuse of notation, we simply denote the loss function $\ell = L \circ h $ with $\ell\in\{0,1\}$.

According to $f$-divergence data processing inequality, if we set the deterministic function $g$ as 
$g(Z) =\mathbf{1}_{E}(Z)$ for any event $E$, then $Y$ is Bernoulli distribution with parameter $P(E)$ or $Q(E)$ and the data processing inequality becomes:
\begin{equation*}
     D_{f}(\mathrm{Bern}(P(E))\|\mathrm{Bern}(Q(E))) \leq D_{f}(P_{Z}\|Q_{Z})
\end{equation*}
If we define the event $E$ as we make an error in the prediction (a.k.a $l(z)=1$), then $P(E) = P(\text{making an error})= E_{P}\mathbf{1}\{\text{making an error}\}=\E_{P}[\ell(z)]$. Therefore we have:
\begin{equation*}
     D_{f}(\mathrm{Bern}(\E_{P}[\ell(z)])\|\mathrm{Bern}(\E_{Q}[\ell(z)])) \leq D_{f}(P_{Z}\|Q_{Z})
\end{equation*}
Again we introduce the intermediate distribution $\calM = \frac{1}{2}(\calS+\calT)$. According to the data processing inequality on the expectation of random variables, if we adopt KL divergence by letting $f(t) =t\log(t)$, then we have:
\begin{equation*}
\begin{split}
    & D_{\text{KL}}(\mathrm{Bern}(\E_{\calT}[\ell(z)])\|\mathrm{Bern}(\E_{\calM}[\ell(z)])) \leq D_{\text{KL}}(\calT\|\calM) \\
    & D_{\text{KL}}(\mathrm{Bern}(\E_{\calS}[\ell(z)])\|\mathrm{Bern}(\E_{\calM}[\ell(z)])) \leq D_{\text{KL}}(\calS\|\calM)
\end{split}
\end{equation*}
We notice $\E_{\calT}(\ell(z))\in[0,1]$, $\E_{\calS}(\ell(z))\in[0,1]$. Then we can adopt Pinsker's inequality by treating the expected value as the Bernoulli distribution parameters. Then we can compute their Total Variation (TV) distance.
\begin{equation*}
    \mathrm{TV}(\mathrm{Bern}(p),\mathrm{Bern}(q))=|p-q|+|1-p-1+q|=2|p-q|
\end{equation*}
Then we have:
\begin{equation*}
\begin{split}
     2|\E_{\calT}[\ell(z)]-\E_{\calM}[\ell(z)]| & = \mathrm{TV}(\mathrm{Bern}(p),\mathrm{Bern}(q))\\
     & \leq \sqrt{2D_{\text{KL}}(\mathrm{Bern}(\E_{\calT}[\ell(z)])\|\mathrm{Bern}(\E_{\calM}[\ell(z)]))}\\
     & \leq \sqrt{2D_{\text{KL}}(\calT\|\calM)}
\end{split}
\end{equation*}
Similarity we have $2|\E_{\calS}[\ell(z)]-\E_{\calM}[\ell(z)]|\leq \sqrt{2D_{\text{KL}}(\calS\|\calM)}$. Adding these two item together we have:
\begin{equation*}
    2|\E_{\calS}[\ell(z)]-\E_{\calM}[\ell(z)]| + 2|\E_{\calT}[\ell(z)]-\E_{\calM}[\ell(z)]| \leq \sqrt{2D_{\text{KL}}(\calT\|\calM)} + \sqrt{2D_{\text{KL}}(\calS\|\calM)}
\end{equation*}
We adopt the inequality $\sqrt{a}+\sqrt{b} \leq \sqrt{2(a +b)}$ with $a\geq 0$ and $b\geq 0$, then we have
\begin{equation*}
    \sqrt{D_{\text{KL}}(\calT\|\calM)} + \sqrt{D_{\text{KL}}(\calS\|\calM)}\leq 2\sqrt{D_{\mathrm{JS}}(\calT\|\calS)}.
\end{equation*}
We also have 
\begin{equation*}
    \begin{split}
       & 2|\E_{\calS}[\ell(z)]-\E_{\calM}[\ell(z)]| + 2|\E_{\calT}[\ell(z)]-\E_{\calM}[\ell(z)]| \\
       & \geq 2|\E_{\calS}[\ell(z)]-\E_{\calM}[\ell(z)]-\E_{\calT}[\ell(z)]+\E_{\calM}[\ell(z)]|\\
       & = 2|\E_{\calS}[\ell(z)]-\E_{\calT}[\ell(z)]|
    \end{split}
\end{equation*}
Given the aforementioned results, we have the following the two side inequality:
\begin{equation*}
    |\E_{\calS}[\ell(z)]-\E_{\calT}[\ell(z)]|\leq \sqrt{D_{\mathrm{JS}}(\calT\|\calS)}
\end{equation*}
We have $-\sqrt{D_{\mathrm{JS}}(\calT\|\calS)}\leq \E_{\calT}[\ell(z)]-\E_{\calS}[\ell(z)]\leq \sqrt{D_{\mathrm{JS}}(\calT\|\calS)}$ and finally we have the lower bound:
\begin{equation*}
    \E_{\calT}[\ell(z)] \geq \E_{\calS}[\ell(z)]  -\sqrt{D_{\mathrm{JS}}(\calT\|\calS)}
\end{equation*}
\paragraph{Remark} We should point out the derived upper bound is looser and restrictive than that we derived from Theorem 1, with a scale $\frac{1}{\sqrt{2}}$ when we restrict the loss in $\{0,1\}$ and Theorem 1 can be extended to any bounded loss while this proof \textbf{cannot}.
\end{proof}

\section{Joint Jensen-Shannon Divergence Decomposition}
In this section, we will provide an upper bound of the chain rule in Jensen-Shannon divergence. 
According to the definition of Jensen-Shannon divergence and the chain rule of KL divergence we have:
\begin{equation*}
    \begin{split}
        2 D_{\text{JS}}(\calT(x,y)\|\calS(x,y)) & =  D_{\text{KL}}(\calT(x,y)\|\calM(x,y)) +  D_{\text{KL}}(\calS(x,y)\|\calM(x,y)) \\
        & = D_{\text{KL}}(\calT(x)\|\calM(x)) + \E_{x\sim\calT(x)} D_{\text{KL}}(\calT(y|x)\|\calM(y|x)) \\
        & + D_{\text{KL}}(\calS(x)\|\calM(x)) + \E_{x\sim\calS(x)} D_{\text{KL}}(\calS(y|x)\|\calM(y|x)) \\
        & = 2 D_{\mathrm{JS}}(\calT(x)\|\calS(x)) + \E_{x\sim\calT(x)} D_{\text{KL}}(\calT(y|x)\|\calM(y|x)) + \E_{x\sim\calS(x)} D_{\text{KL}}(\calS(y|x)\|\calM(y|x)) 
    \end{split}
\end{equation*}
In general, for continuous random variable, the $D_{\text{KL}}$ divergence does not exist an exact upper bound. While we can simple upper bound these by adding two complementary terms.
\begin{equation*}
\begin{split}
     \E_{x\sim\calT(x)} D_{\text{KL}}(\calT(y|x)\|\calM(y|x)) & \leq \E_{x\sim\calT(x)} D_{\text{KL}}(\calT(y|x)\|\calM(y|x)) + \E_{x\sim\calT(x)} D_{\text{KL}}(\calS(y|x)\|\calM(y|x)) \\
    & = 2 \E_{x\sim\calT(x)} D_{\mathrm{JS}}(\calT(y|x)\|\calS(y|x))
\end{split}
\end{equation*}

\begin{equation*}
\begin{split}
     \E_{x\sim\calS(x)} D_{\text{KL}}(\calT(y|x)\|\calM(y|x)) & \leq \E_{x\sim\calS(x)} D_{\text{KL}}(\calT(y|x)\|\calM(y|x)) + \E_{x\sim\calS(x)} D_{\text{KL}}(\calS(y|x)\|\calM(y|x)) \\
    & = 2 \E_{x\sim\calS(x)} D_{\mathrm{JS}}(\calT(y|x)\|\calS(y|x))
\end{split}
\end{equation*}

Plugging in the results, we have the following conditional upper bound
\begin{equation*}
 D_{\mathrm{JS}}(\calT(x,y)\|\calS(x,y)) \leq   D_{\mathrm{JS}}(\calT(x)\|\calS(x)) + 
 \E_{x\sim\calT(x)} D_{\mathrm{JS}}(\calT(y|x)\|\calS(y|x)) + \E_{x\sim\calS(x)} D_{\mathrm{JS}}(\calT(y|x)\|\calS(y|x))
\end{equation*}

We can derive the analogue result conditioned on $y$:
\begin{equation*}
 D_{\mathrm{JS}}(\calT(x,y)\|\calS(x,y)) \leq   D_{\mathrm{JS}}(\calT(y)\|\calS(y)) + 
 \E_{y\sim\calT(y)} D_{\mathrm{JS}}(\calT(x|y)\|\calS(x|y)) + \E_{y\sim\calS(y)} D_{\mathrm{JS}}(\calT(x|y)\|\calS(x|y))
\end{equation*}

\section{Target Intrinsic Error Upper Bound}

\begin{tcolorbox}
If $H(Y_s|X_s)\leq \epsilon$, the source target marginal and conditional distribution
are close $D_{\text{JS}}(\calS(x)\|\calT(x))\leq\delta_1$, $\forall x$, we have $D_{\text{JS}}(\calS(y|x)\|\calT(y|x))\leq\delta_2$. Then the target distribution conditional entropy can be upper bounded by:
\begin{equation*}
    H(Y_t|X_t)\leq \epsilon + \sqrt{\frac{\delta_2}{2}} + \frac{\sqrt{\delta_1}}{2}\log|\calY|
\end{equation*}
\end{tcolorbox}

\begin{proof}
Since $\frac{1}{2}\text{TV}(P,Q)^2\leq D_{\text{JS}}(P\|Q)\leq TV(P,Q)$ \cite{thekumparampil2018robustness}, then 
for $\forall x$ we have: 
\begin{equation*}
    \|\calS(y|x)-\calT(y|x)\|_1 \leq \sqrt{2\delta_2}
\end{equation*}
Then for conditional entropy for the target distribution, we have:
\begin{equation*}
\begin{split}
    H(Y_t|X_t) & = \E_{x\sim\calT(x)} H(Y_t|X_t=x) \\
                & = \E_{x\sim\calT(x)} H(Y_t|X=x) - \E_{x\sim\calT(x)} H(Y_s|X=x) + \E_{x\sim\calT(x)} H(Y_s|X=x)\\
                & \leq \E_{x\sim\calT(x)} |H(Y_t|X=x)-H(Y_s|X=x)| + \E_{x\sim\calT(x)} H(Y_s|X=x)
\end{split}
\end{equation*}
Since the Entropy function is $\frac{1}{2}$ Lipschitz w.r.t. $L_1$ norm, then we have
\begin{equation*}
    \E_{x\sim\calT(x)} |H(Y_t|X=x)-H(Y_s|X=x)| \leq \E_{x\sim\calT(x)} \frac{1}{2}\|\calT(y|x)-\calS(y|x) \|_1 \leq \sqrt{\frac{\delta_2}{2}}
\end{equation*}
Then we need to bound $\E_{x\sim\calT(x)} H(Y_s|X=x)$, 
\begin{equation*}
\begin{split}
    E_{x\sim\calT(x)} H(Y_s|X=x) & = E_{x\sim\calS(x)} H(Y_s|X=x) + E_{x\sim\calT(x)} H(Y_s|X=x) - E_{x\sim\calS(x)} H(Y_s|X=x) \\
                                 & \leq \epsilon + E_{x\sim\calT(x)} H(Y_s|X=x) - E_{x\sim\calS(x)} H(Y_s|X=x)
\end{split}
\end{equation*}

We still adopt the conclusion when we proof Theorem 1, i.e the transport inequality of the gaps of same function under different marginal distribution measures by assuming $z=x$.
We can compute $G = H(Y_s|X=x)\leq H(Y_s) \leq \log|\calY|$, then we have:
\begin{equation*}
    \begin{split}
        E_{x\sim\calT(x)} H(Y_s|X=x) & \leq \epsilon + \frac{\log|\calY|}{\sqrt{2}} \sqrt{D_{\text{JS}}(\calT(x)\|\calS(x))} \\
                                     & \leq \epsilon + \sqrt{\frac{\delta_1}{2}} \log|\calY|
    \end{split}
\end{equation*}
Putting all them together we have the aforementioned conclusion.
\end{proof}

\section{Inherent Difficulty for Controlling Label Conditional Shift}

\subsection{Extension to the Representation Learning}
\begin{tcolorbox}
The upper bound in Theorem 1 can be further decomposed as:
\begin{equation}
\begin{split}
      R_{\calT}(h) \leq & R_{\calS}(h) + \frac{G}{\sqrt{2}}\sqrt{D_{\mathrm{JS}}(\calT(x)\|\calS(x))} \\
      & + \frac{G}{\sqrt{2}}\sqrt{\E_{x\sim\calT(x)} D_{\mathrm{JS}}(\calT(\cdot|x)\|\calS(\cdot|x)) + \E_{x\sim\calS(x)} D_{\mathrm{JS}}(\calT(\cdot|x)\|\calS(\cdot|x))}
\end{split}
\end{equation}
\end{tcolorbox}

Inspired by \cite{johansson2020generalization}, we set the representation function $g:\calX\to\mathcal{Z}$ and $h$ the hypothesis defined on the $(x,z)$. Then we consider learning twice-differentiable, invertible representations: $g:\calX\to\mathcal{Z}$ where $g^{-1}$ is the inverse representation, such that $g^{-1}(g(x))=x$ for all $x$. Then these assumptions for $g(x)$, we have $P(g(X)=z) = P(X=g^{-1}(z))$. 

\begin{tcolorbox}
We can therefore extend the result in the representaion learning:
\begin{equation*}
     \begin{split}
      R_{\calT}(h\circ g) \leq & R_{\calS}(h\circ g) + \frac{G\sqrt{A_{g}}}{\sqrt{2}}\sqrt{D_{\mathrm{JS}}(\calT(z)\|\calS(z))} \\
      & + \frac{G}{\sqrt{2}}\sqrt{\E_{x\sim\calT(x)} D_{\mathrm{JS}}(\calT(\cdot|z)\|\calS(\cdot|z)) + \E_{x\sim\calS(x)} D_{\mathrm{JS}}(\calT(\cdot|z)\|\calS(\cdot|z))}
\end{split}
\end{equation*}
Where $A(g)=\sup_{z}|J_{g^{-1}}(z)|$, is the maximum value of the Jacobian of the representation inverse function $g^{-1}$.
\end{tcolorbox}

As we mentioned in this and previous paper \cite{wu2020representation,zhao2019learning,johansson2019support,wu2019domain}, only controlling the first two terms by learning a bad representation can lead to the third term much larger.

\begin{proof}
According to the definition of $f$-divergence and define $z=g(x)$, under the aforementioned assumptions, we have:
\begin{equation*}
\begin{split}
    D_{f}(P(x)\|Q(x)) & =  \int_{x} q(x)f(\frac{p(x)}{q(x)})dx \\
    & = \int_{z} q(g(x))f(\frac{p(g(x))}{q(g(x))}) |J_{g^{-1}}(z)|dz\\
    & \leq A_{g}  D_{f}(P(z)\|Q(z))
\end{split}
\end{equation*}
Where $A(g)=\sup_{z}|J_{g^{-1}}(z)|$, is the maximum value of the Jacobian of the representation inverse function $g^{-1}$. 
\end{proof}

\subsection{Non-Asymptotic Analysis}
Based on the standard statistical learning theory method, we can further derive the non-asymptotic bound. According to \cite{mohri2018foundations}, the empirical risk can finally converge to its expected counterpart, informally $\forall h\in\mathcal{H}, g\in\mathcal{G}$ with high probability we have:
\begin{equation}
    R_{\calS}(h \circ g) \leq \hat{R}_{\calS}(h \circ g) + \mathcal{O}(\frac{1}{\sqrt{N_S}})
\end{equation}

As for estimation the empirical marginal distribution from the data, according to \cite{biau2018some}, informally estimating empirical Jensen-Shannon divergence satisfies the standard convergence rate $\mathcal{O}(\frac{1}{\sqrt{N_S}}+\frac{1}{\sqrt{N_T}})$.
\begin{equation}
     D_{\text{JS}}(\calT(g(x)\| \calS(g(x)) \leq D_{\text{JS}}(\hat{\calT}(g(x)\| \hat{\calS}(g(x))  +  \mathcal{O}(\frac{1}{\sqrt{N_S}} + \frac{1}{\sqrt{N_T}})
\end{equation}

\subsection{Lower Bound of Label Conditional Shift}
\begin{tcolorbox}[grow to left by=0.3cm,grow to right by=0.3cm]
We can prove the label-conditional shift can be lower bounded by:
\begin{equation*}
\begin{split}
      \E_{z\sim\hat{\calT}(z)} D_{\text{JS}}(\hat{\calS}(y|z)\| \hat{\calT}(y|z)) & + \E_{z\sim\hat{\calS}(z)} D_{\text{JS}}(\hat{\calS}(y|z)\| \hat{\calT}(y|z)) \\
      & \geq 2\left(\sqrt{D_{\text{JS}}(\hat{\calT}(y)\|\hat{\calS}(y))} - \sqrt{D_{\text{JS}}(\hat{\calS}(z)\| \hat{\calT}(z))} \right)^2
      \end{split}
\end{equation*}
\end{tcolorbox}

We notice the square form of Jensen-Shannon divergence is the valid statistical distance. Then we have:
\begin{equation*}
    \begin{split}
       & \sqrt{D_{\text{JS}}(\hat{\calT}(y)\|\hat{\calS}(y))}  = \sqrt{D_{\text{JS}}(\sum_{z} \hat{\calT}(y|z)\hat{\calT}(z)\|\sum_{z} \hat{\calS}(y|z)\hat{\calS}(z))} \\
        & \leq \sqrt{D_{\text{JS}}(\sum_{z} \hat{\calT}(y|z)\hat{\calS}(z)\|\sum_{z} \hat{\calS}(y|z)\hat{\calS}(z))} 
        + \sqrt{D_{\text{JS}}(\sum_{z} \hat{\calT}(y|z)\hat{\calS}(z)\|\sum_{z} \hat{\calT}(y|z)\hat{\calT}(z))} \\
        & \leq  \sqrt{\E_{z\sim\hat{\calS}(z)} D_{\text{JS}}(\hat{\calS}(y|z)\| \hat{\calT}(y|z))} + \sqrt{D_{\text{JS}}(\hat{\calS}(z)\| \hat{\calT}(z))} 
    \end{split}
\end{equation*}
We derive the inequality according to (1) Jensen-Shannon distance is a valid statistical metric; (2) The convex property of the Jensen-Shannon divergence w.r.t. the empirical distribution; (3) The $f$-divergence data-processing inequality.

\begin{equation*}
    \E_{z\sim\hat{\calS}(z)} D_{\text{JS}}(\hat{\calS}(y|z)\| \hat{\calT}(y|z)) \geq 
    \left(\sqrt{D_{\text{JS}}(\hat{\calT}(y)\|\hat{\calS}(y))} - \sqrt{D_{\text{JS}}(\hat{\calS}(z)\| \hat{\calT}(z))} \right)^2
\end{equation*}

We can analogue derive:
\begin{equation*}
    \E_{z\sim\hat{\calT}(z)} D_{\text{JS}}(\hat{\calS}(y|z)\| \hat{\calT}(y|z)) \geq 
    \left(\sqrt{D_{\text{JS}}(\hat{\calT}(y)\|\hat{\calS}(y))} - \sqrt{D_{\text{JS}}(\hat{\calS}(z)\| \hat{\calT}(z))} \right)^2
\end{equation*}

Finally the third term can be lower bounded by:
\begin{equation*}
    \E_{z\sim\hat{\calT}(z)} D_{\text{JS}}(\hat{\calS}(y|z)\| \hat{\calT}(y|z)) + \E_{z\sim\hat{\calS}(z)} D_{\text{JS}}(\hat{\calS}(y|z)\| \hat{\calT}(y|z)) \geq 
    2\left(\sqrt{D_{\text{JS}}(\hat{\calT}(y)\|\hat{\calS}(y))} - \sqrt{D_{\text{JS}}(\hat{\calS}(z)\| \hat{\calT}(z))} \right)^2
\end{equation*}
Which exactly recovers the result of \cite{zhao2019learning}: over-matching the marginal distribution divergence to zero can increase this lower bound of the third term.

\section{New Practical Principles}
\begin{equation}
\begin{split}
      R_{\calT}(h) \leq & R_{\calS}(h) + \frac{G}{\sqrt{2}}\underbrace{\sqrt{D_{\mathrm{JS}}(\calT(y)\|\calS(y))}}_{\text{Label Marginal Shift}} \\
      & + \frac{G}{\sqrt{2}}\underbrace{\sqrt{\E_{y\sim\calT(y)} D_{\mathrm{JS}}(\calT(x|y)\|\calS(x|y)) + \E_{y\sim\calS(y)} D_{\mathrm{JS}}(\calT(x|y)\|\calS(x|y))}}_{\text{Semantic (Cofeature) Conditional Shift}}
\end{split}
\end{equation}

In this section, we firstly prove the lower bound in context of conditional distribution matching. We demonstrate that in the presence of conditional distribution matching, we still need to control the label shift term to control a small lower bound. 

\subsection{Necessity of Considering Label Shift}
In this section, we suppose there exist a more general stochastic representation learning function $g$ with a conditional probability distribution $g(z|x)$. \footnote{The deterministic representation learning function can be viewed as a special case such that fixed $g(z|x)=z$ for a given $x$} Then the marginal distribution and conditional distribution w.r.t. latent variable can be reformulated as:
\begin{equation*}
    \calS(z) = \int_{x} g(z|x) \calS(x) dx \quad\quad  \calS(z|y) = \int_{x} g(z|x) \calS(x|Y=y) dx
\end{equation*}

\begin{tcolorbox}
If $\forall$ classifier $h$, feature function $g$, and label $y\in\calY=\{-1,+1\}$ such that semantic conditional distribution is matched: $D_{\text{JS}}(\calS(z|y),\calT(z|y))=0$, then the target risk can be bounded:
\begin{equation}
      R_{\calS}(h \circ g)- \sqrt{2 D_{\text{JS}}(\calS(y),\calT(y))}  \leq R_{\calT}(h \circ g) \leq R_{\calS}(h\circ g) + \sqrt{2 D_{\text{JS}}(\calS(y),\calT(y))}
\label{eq:representation_bound}
\end{equation}
Where $R_{\calS}(h \circ g) = R_{\calS}(h(g(x),y))$ the expected risk over the classifier $h$ and feature learner $g$.
\end{tcolorbox}

\begin{proof}
For simplifying the analysis, we only focus on the binary classification with margin style loss 
with $L(h(z),y) = L(yh(z))$, including $0-1$ loss, hinge loss, logistic loss, etc). Throughout the whole analysis, we will simply adopt the $0-1$ loss. We additionally define the following distributions:

\begin{equation*}
    \begin{split}
        & \mu^{\calS}(z) = \calS(Y=1,Z=z)  = \calS(Y=1)  \calS(Z=z|Y=1)  \\
        & \pi^{\calS}(z) = \calS(Y=-1,Z=z) = \calS(Y=-1) \calS(Z=z|Y=-1)  \\
        & \mu^{\calT}(z) = \calT(Y=1,Z=z)  = \calT(Y=1)  \calT(Z=z|Y=1)  \\
        & \pi^{\calT}(z) = \calT(Y=-1,Z=z) = \calT(Y=-1) \calT(Z=z|Y=-1)
    \end{split}
\end{equation*}

Then in the source distribution and target distribution for the common feature extractor $Q$ and hypothesis $h$, we have:
$$ R_{\calS}(h\circ g) = \E_{\calS} \mathbf{1}\{yh(z)\leq 0\}$$
$$ R_{\calT}(h\circ g) = \E_{\calT} \mathbf{1}\{yh(z)\leq 0\}$$

According to \cite{nguyen2009surrogate}, the risk can be reformulated as 
\begin{equation*}
\begin{split}
    & R_{\calS}(h\circ g) = \sum_{z} \mathbf{1}\{h(z)\leq 0\} \mu^{\calS}(z) +   \mathbf{1}\{h(z)>0\}\pi^{\calS}(z) \\
    &  R_{\calT}(h\circ g) = \sum_{z} \mathbf{1}\{h(z)\leq 0\} \mu^{\calT}(z) +   \mathbf{1}\{h(z)>0\}\pi^{\calT}(z)
\end{split}
\end{equation*}
Then we have:
\begin{equation*}
\begin{split}
     R_{\calT}(h\circ g) - R_{\calS}(h\circ g) 
     & =  \sum_{z}  \mathbf{1}\{h(z)\leq 0\} \left(\mu^{\calT}(z)- \mu^{\calS}(z)\right) + \mathbf{1}\{h(z)>0\} (\pi^{\calT}(z)-\pi^{\calS}(z)) \\
     & \geq \sum_{z} \min\{\mu^{\calT}(z)- \mu^{\calS}(z), \pi^{\calT}(z)-\pi^{\calS}(z)\}
\end{split}
\end{equation*}

If we define the conditional distribution matching as there exists a distribution $\exists g$ such that $\calS(z|y) = \calT(z|y)=\D(z|y)$, then we can simplify as

\begin{equation*}
\begin{split}
     & \sum_{z} \min\{\mu^{\calT}(z)- \mu^{\calS}(z), \pi^{\calT}(z)-\pi^{\calS}(z)\} \\
     & \geq -|\calS(y=1)-\calT(y=1)|\sum_{z} \max \{\D(z|y=1),\D(z|y=-1)\}\\
     & = -\frac{1}{2} d_{\text{TV}}(\calS(y),\calT(y)) \frac{1}{2}(1+d_{\text{TV}}(\D(z|y=1),\D(z|y=-1))\\
     & \geq -\frac{1}{2} d_{\text{TV}}(\calS(y),\calT(y))\frac{1}{2}(1+1)= -\frac{1}{2} d_{\text{TV}}(\calS(y),\calT(y))\geq -\sqrt{2 D_{\text{JS}}(\calS(y),\calT(y))}
\end{split}
\end{equation*}

As for the upper bound, since we have:
\begin{equation*}
\begin{split}
     R_{\calT}(h\circ g) - R_{\calS}(h \circ g) 
     & =  \sum_{z}  \mathbf{1}\{h(z)\leq 0\} \left(\mu^{\calT}(z)- \mu^{\calS}(z)\right) + \mathbf{1}\{h(z)>0\} (\pi^{\calT}(z)-\pi^{\calS}(z)) \\
     & \leq \sum_{z} \max\{\mu^{\calT}(z)- \mu^{\calS}(z), \pi^{\calT}(z)-\pi^{\calS}(z)\}
\end{split}
\end{equation*}
Given the conditional shift, we have:
\begin{equation*}
\begin{split}
     & \sum_{z} \max\{\mu^{\calT}(z)- \mu^{\calS}(z), \pi^{\calT}(z)-\pi^{\calS}(z)\} \\
     & \leq |\calS(y=1)-\calT(y=1)|\sum_{z} \max \{\D(z|y=1),\D(z|y=-1)\}\\
     & = \frac{1}{2} d_{\text{TV}}(\calS(y),\calT(y)) \frac{1}{2}(1+d_{\text{TV}}(\D(z|y=1),\D(z|y=-1))) \\
     & \leq \frac{1}{2} d_{\text{TV}}(\calS(y),\calT(y))\frac{1}{2}(1+1)= \frac{1}{2} d_{\text{TV}}(\calS(y),\calT(y))\leq \sqrt{2 D_{\text{JS}}(\calS(y),\calT(y))}
\end{split}
\end{equation*}

Finally we have the two side bound:
\begin{equation*}
 R_{\calS}(h\circ g)- \sqrt{2 D_{\text{JS}}(\calS(y),\calT(y))}  \leq R_{\calT}(h\circ g) \leq R_{\calS}(h\circ g) + \sqrt{2 D_{\text{JS}}(\calS(y),\calT(y))}
\end{equation*}
\end{proof}

\subsection{Labeling Shift Correction: Theoretical Result}\label{sec:label_shift}
As our previous theoretical results indicate the necessarily of label shift correction. 
If the semantic (cofeature) conditional distribution is matched  $D_{\text{JS}}(\calT(z|y)\|\calS(z|y))=0$, we adopt the popular label re-weighted loss strategy: $\hat{R}^{\alpha}_{\calS}(h \circ g) = \sum_{(x_s,y_s)\in\hat{\calS}} \alpha(y_s) L(h(g(x_s),y_s))$ with $\alpha(y) = \frac{\calT(y)}{\calS(y)}$. Then for $\forall h\in\calH$ with high probability, we have:
\begin{equation*}
    |\hat{R}^{\alpha}_{\calS}(h)-R_{\calT}(h)|\leq  \mathcal{O}(\sqrt{\frac{D_{\text{JS}}(\calT(y)\|\calS(y))}{N_S}})
\end{equation*}

\begin{proof}
According to the Lemma 4 of \cite{azizzadenesheli2018regularized}, for a given hypothesis class $\calH$, under $N$ data points we have:
\begin{align*}
    \sup_{h\in\calH} |\hat{R}^{\alpha}_{\calS}(h) - R_{\calT}(h)|\leq \mathcal{O}(\sqrt{\frac{d_2(\calT(y)\|\calS(y))\log(2/\delta)}{N}})
\end{align*}
with probability at least $1-\delta$. 

Since $d_2(\calT(y)\|\calS(y)) = 2^{D_2\left(\calT(y)\|\calS(y)\right)}$ with $D_2\left(\calT(y)\|\calS(y)\right)$ is the R\'enyi-2 divergence. Then according to the \cite{sason2015upper,thekumparampil2018robustness}, there exists a positive constant $C^{\prime}$ such that: 
\begin{equation*}
    D_2(\calT(y)\|\calS(y)) \leq \log(1+C^{\prime}d_{TV}(\calT(y),\calS(y))) \leq \log(1+C^{\prime}D_{\text{JS}}(\calT(y)\|\calS(y)))
\end{equation*}
Plugging in the model, we have:
\begin{equation*}
    \sup_{h\in\calH} |\hat{R}^{\alpha}_{\calS}(h) - R_{\calT}(h)|\leq \mathcal{O}(\sqrt{\frac{D_{\text{JS}}(\calT(y)\|\calS(y))}{N_S}})
\end{equation*}
\end{proof}

\subsection{Detecting Poor Pseudo-Label}
\begin{tcolorbox}
We can prove if we have poor pseudo-label, the marginal divergence can be very large.
If we assume $D_{\text{JS}}(\hat{\calT}(y)\|\hat{\calT}_{p}(y)) = P $, and small source prediction error $D_{\text{JS}}(\hat{\calS}(y)\|\hat{\calS}_{p}(y))\leq \epsilon_1$ and small source target ground truth distribution   $D_{\text{JS}}(\hat{\calS}(y)\|\hat{\calT}(y))\leq \epsilon_2$, 
then we can prove 
\begin{equation*}
    D_{\text{JS}}(\hat{\calS}(z)\|\hat{\calT}(z))\geq (\sqrt{P}-\sqrt{\epsilon_1}-\sqrt{\epsilon_2})^2
\end{equation*}
\end{tcolorbox}

\begin{proof}
Since in the DA, we adopt the same classifier $h$ to predict both domains, the empirical label prediction output distribution (pseudo-label distribution) is defined as:

\begin{equation*}
      \hat{\calS}_p(y) = \sum_{z} h(y|z)\hat{\calS}(z) \quad\quad
      \hat{\calT}_p(y) = \sum_{z} h(y|z)\hat{\calT}(z)
\end{equation*}

According to the $f$-divergence data-processing inequality, we have:
$$ D_{\text{JS}}(\hat{\calS}(z)\|\hat{\calT}(z)) \geq  D_{\text{JS}}(\hat{\calS}_p(y)\|\hat{\calT}_p(y)) $$

Since Jensen-Shannon distance is a valid statistical distance, then we have:
\begin{equation*}
  \sqrt{D_{\text{JS}}(\hat{\calS}_p(y)\|\hat{\calT}_p(y))} +   \sqrt{D_{\text{JS}}(\hat{\calS}_p(y)\|\hat{\calS}(y))} + \sqrt{D_{\text{JS}}(\hat{\calS}(y)\|\hat{\calT}(y))} \geq \sqrt{D_{\text{JS}}(\hat{\calT}(y)\|\hat{\calT}_{p}(y))} =\sqrt{P}
\end{equation*}
Since we have a small source prediction error, a small empirical label shift, then we have:
\begin{equation*}
    \sqrt{D_{\text{JS}}(\hat{\calS}_p(y)\|\hat{\calT}_p(y))} \geq \sqrt{P}-\sqrt{\epsilon_1}-\sqrt{\epsilon_2}
\end{equation*}
Combining together we have $D_{\text{JS}}(\hat{\calS}(z)\|\hat{\calT}(z))\geq (\sqrt{P}-\sqrt{\epsilon_1}-\sqrt{\epsilon_2})^2$
\end{proof}

\section{Practical Guidelines}
\begin{tcolorbox}[grow to left by=0.3cm,grow to right by=0.3cm]
\textbf{Parameter Optimization Step} (fixed Pseudo-Labels) classifier $h$ and feature extractor $g$:
\begin{equation*}
    \begin{split}
         & \min_{h,g}\quad \underbrace{\hat{R}^{\hat{\alpha}}_{\calS}(h(g(x_s),y_s))}_{(\RN{1})} + \underbrace{\sum_{y} (\hat{\calS}(y) + \hat{\calT}_p(y))D_{\text{JS}}\left(\hat{\calT}(g(x_t)|Y_p=y)\|\hat{\calS}(g(x_s)|Y=y)\right)}_{(\RN{2})} \\
         & \text{s.t.}\quad\underbrace{D_{\text{JS}}(\hat{\calT}(g(x_t))\|\hat{\calS}(g(x_s)))\leq \kappa}_{(\RN{3})}
    \end{split}
\end{equation*}
$(\RN{1})$~Labeling shift correction:$\hat{R}^{\hat{\alpha}}_{\calS}(h(g(x_s),y_s)) = \frac{1}{N_S}\sum_{i=1}^{N_S} \alpha(y_i) L(h(g(x_i),y_i))$; $(\RN{2})$~Semantic conditional  matching, to align the semantic feature; $(\RN{3})$~Cofeature marginal distribution matching as the constraint, as a board adaptation step to obtain a good initialization pseudo-label prediction.
\end{tcolorbox}
\begin{tcolorbox}[grow to left by=0.3cm,grow to right by=0.3cm]
\textbf{Pseudo-Label Estimation Step} (fixed Parameters):  \quad\quad\quad  $y^{p}$, $\hat{\alpha}$, $\hat{\calT}_p(y)$ \\
$y^{p}$,$\hat{\calT}_p(y)$ are pseudo-labels and distributions on the target domain. $\hat{\alpha}$ label reweighting coefficient. 
\end{tcolorbox}

\subsection{Semantic Conditional Distribution Matching (Principle $\RN{2}$)}
As we illustrated in the paper, the first component is to match the cofeature conditional distribution divergence. Then we have:
\begin{equation}
\begin{split}
    \sum_{y} (\hat{\calS}(y) + \hat{\calT_p}(y)) D_{\text{JS}}(\hat{\calT}(\cdot|y)\|\hat{\calS}(\cdot|y)) & \leq \sum_{y} (\hat{\calS}(y) + \hat{\calT_p}(y)) d_\text{TV}(\hat{\calT}(\cdot|y)\|\hat{\calS}(\cdot|y)) \\
    & \leq C \sum_{y} (\hat{\calS}(y) + \hat{\calT_p}(y)) \|\hat{\calT}(\cdot|y)-\hat{\calS}(\cdot|y) \|_2 
\end{split}
    \label{eq:y_cond_upper_2}
\end{equation}
In the representation learning, we simply approximate the empirical distribution as the surrogate of the conditional distribution. We therefore denote:
\begin{equation*}
    \hat{\calS}(g(x_s)|y)\approx \frac{1}{|\# y_s=y|}\sum_{(x_s,y_s)} \delta_{\{y_s=y\}} g(x_s) \quad\quad\quad
    \hat{\calT}(g(x_t)|y)\approx \frac{1}{|\# y^{p}_t=y|}\sum_{(x_t,y_t^{p})} \delta_{\{y_t^{p}=y\}} g(x_t) 
\end{equation*}
Therefore the conditional matching term can be approximated as:
\begin{equation}
    \hat{R}_{\text{cond}}(g) =  \sum_{y} (\hat{\calS}(Y=y) + \hat{\calT}_{p}(Y=y)) \|\hat{\calS}(g(x_s)|Y=y)-\hat{\calT}(g(x_t)|Y_p=y)\|_2^2 
\end{equation}
\paragraph{Remark} We would like to emphasize that we propose one feasible solution. The cofeature conditional distribution matching can be naturally extended to the conditional adversarial training \cite{long2018conditional}, matching higher statistical moments \cite{cai2019learning} or infinite orders as MMD distance \cite{long2015learning}.

\subsection{Marginal Cofeature Distribution Matching as the Constraint (Principle $\RN{3}$)}
Since the a relative accurate pseudo-label estimation is important in the iterative algorithm, thus we introduce the marginal cofeature distribution matching as the training constraint. The main goal is to keep a good pseudo-label initial estimation. We just adopt the most popular Jensen-Shannon domain adversarial training (the dual term of linear shift Jensen-Shannon divergence)
\begin{equation}
     \hat{R}_{\text{adv}}(d,g) = \E_{x_s\sim\hat{\calS}(x)} \log(d \circ g (x_s)) +  \E_{x_t\sim\hat{\calT}(x)} \log(1- d \circ g (x_t))
\end{equation}

As for the constraints, we adopt Lagrangian relaxation approach as treat the constraint as a small regularization term, where $\kappa$ is the hyper-parameter.

\subsection{Labeling Marginal Shift Correction (Principle $\RN{1}$)}
We adopt the cross entropy as classification loss, then we have:
\begin{equation}
    \hat{R}^{\hat{\alpha}}_{\calS}(f,g) = - \frac{1}{N_S}\sum_{(x_s,y_s)\sim\hat{\calS}} \hat{\alpha}(y_s) \log(h\circ (g(x_s),y_s))
\end{equation}

\paragraph{Estimation $\hat{\alpha}$ and Target label distribution}
We follow the popular (Black Box Shift Learning) BBSL estimator. We first construct a source prediction confusion matrix $\hat{C}\in|\calY|\times|\calY|$ with $\hat{C}[i,j] = \Proba(\mathrm{argmax}_{y}~h(g(x_s),y)=i, y_s=j)$. The target pseudo-label $y^p$ and target pseudo-label distribution $\hat{\calT}_p$ can be directly estimated from the neural network. Then the label re-weighting coefficient can be estimated as:
$$\hat{\alpha} = \hat{C}^{-1}\hat{\calT}_p $$

\subsection{Practical Loss}
We consider the whole aforementioned components and derive the following training strategy.
\begin{tcolorbox}
\emph{Parameter Optimization}
\begin{equation*}
    \min_{f,g} \max_{d} \hat{R}(f,d,g) = \hat{R}^{\hat{\alpha}}_{\calS}(f,g) +  \lambda_0 \hat{R}_{\text{adv}}(d,g) + \lambda_1 \hat{R}_{\text{cond}}(g)
\end{equation*}
\end{tcolorbox}

\begin{tcolorbox}
\emph{Pseudo-Label Estimation}
\begin{equation*}
    \hat{\alpha} = \hat{C}^{-1}\hat{\calT}_p
\end{equation*}
The source confusion matrix $\hat{C}$, target pseudo-label $y^p$ and target pseudo-label distribution $\hat{\calT}_p$ can be directly estimated from the neural network.
\end{tcolorbox}

\section{Experimental Descriptions (Sec.~6)}
\subsection{Dataset Descriptions}
\paragraph{Office-31} \cite{saenko2010adapting} This dataset is widely used for visual domain adaptation. It consists of 4,652 images and 31 categories collected from three different domains: Amazon (A) from amazon.com, Webcam (W) and DSLR (D), taken by web camera and digital SLR camera in different environmental settings, respectively. We test all the domain combinations.

\paragraph{ImageCLEF} \cite{villegas2015general} This data is originally used for the ImageCLEF 2014 domain adaptation challenge consists of twelve common classes from three domains: ImageNet ILSVRC 2012 (I), Pascal VOC 2012 (P), and Caltech-256 (C). Each doamin has 600 images in total. We test 6 tasks by using all domain combinations.

\subsection{Experimental Details}
We finetune the AlexNet pre-trained network from the ImageNet. Following the Domain Adversarial Neural Network, we gradually increase the weight of $\hat{R}_{\text{adv}}$ by setting $\lambda_0 = \frac{2}{1+\exp(km)}-1$, where $k=-10$ and $m$ denotes the training progress from $0$ to $1$. We set $\lambda_1 = K \lambda_0$ ($K>1$) to given a higher weight for semantic conditional distribution matching, during the training. For optimizing the semantic conditional loss, we use the moving average strategy to estimate its centroid, the same in \cite{xie2018learning}.

\subsection{Comparison Methods}
We compare the methods which merely considered one part, or two principles and demonstrate their performances.
\begin{enumerate}
    \item DANN \cite{ganin2016domain}, approach merely consider the principle $(\RN{3})$ restricting a small marginal Jensen-Shannon divergence;
    \item CDAN \cite{long2018conditional}, approach merely \textbf{implicitly} consider the principle $(\RN{2})$, minimize the conditional shift;
    \item Principles W.o $(\RN{3})$ (without marginal distribution matching constraint)
    \item Principles W.o $(\RN{1})$ (without label marginal shift correction)
    \item Principles W.o $(\RN{2})$ (without semantic conditional matching)
\end{enumerate}

\end{document}